\DocumentMetadata{
  pdfversion=2.0,pdfstandard=ua-2,
  testphase={phase-IV,firstaid,math,title}
}

\documentclass[sigconf,screen]{acmart-tagged}
\usepackage{multirow}    
\usepackage{tabularx}    
\usepackage{ragged2e}    
\usepackage{amsfonts}    
\usepackage{makecell}
\usepackage{microtype}

\AtBeginDocument{
  }

\copyrightyear{2026}
\acmYear{2026}
\setcopyright{cc}
\setcctype{by}
\acmConference[HRI '26]{Proceedings of the 21st ACM/IEEE International Conference on Human-Robot Interaction}{March 16--19, 2026}{Edinburgh, Scotland UK}
\acmBooktitle{Proceedings of the 21st ACM/IEEE International Conference on Human-Robot Interaction (HRI '26), March 16--19, 2026, Edinburgh, Scotland UK}
\acmPrice{}
\acmDOI{10.1145/3757279.3785617}
\acmISBN{979-8-4007-2128-1/2026/03}

\begin{document}

\title[Explaining Why Things Go Where They Go...]{Explaining Why Things Go Where They Go: Interpretable Constructs of Human Organizational Preferences}

\author{Emmanuel Fashae}
\affiliation{%
  \institution{Monash University, CSIRO Robotics}
  \city{Melbourne}
  \country{Australia}
}

\email{emmanuel.fashae@monash.edu}

\author{Michael Burke}
\affiliation{%
  \institution{Monash University}
  \city{Melbourne}
  \country{Australia}
}
\email{michael.g.burke@monash.edu}

\author{Leimin Tian}
\affiliation{%
  \institution{CSIRO Robotics}
  \city{Melbourne}
  \country{Australia}
}

\email{leimin.tian@csiro.au}

\author{Lingheng Meng}
\affiliation{%
  \institution{CSIRO Robotics}
  \city{Melbourne}
  \country{Australia}
}
\email{lingheng.meng@csiro.au}

\author{Pamela Carreno-Medrano}
\affiliation{%
  \institution{Monash University}
  \city{Melbourne}
  \country{Australia}
}
\email{pamela.carreno@monash.edu}

\renewcommand{\shortauthors}{E.Fashae, M.Burke, L.Tian, L.Meng, and P.Carreno-Medrano}

\begin{abstract}
Robotic systems for household object rearrangement often rely on latent preference models inferred from human demonstrations. While effective at prediction, these models offer limited insight into the interpretable factors that guide human decisions. We introduce an explicit formulation of object arrangement preferences along four interpretable constructs: spatial practicality (putting items where they naturally fit best in the space), habitual convenience (making frequently used items easy to reach), semantic coherence (placing items together if they are used for the same task or are contextually related), and commonsense appropriateness (putting things where people would usually expect to find them). To capture these constructs, we designed and validated a self-report questionnaire through a 63-participant online study. Results confirm the psychological distinctiveness of these constructs and their explanatory power across two scenarios (kitchen and living room). We demonstrate the utility of these constructs by integrating them into a Monte Carlo Tree Search (MCTS) planner and show that when guided by participant-derived preferences, our planner can generate reasonable arrangements that closely align with those generated by participants. This work contributes a compact, interpretable formulation of object arrangement preferences and a demonstration of how it can be operationalized for robot planning.
\end{abstract}

\ccsdesc[500]{Computer systems organization~Robotics}
\ccsdesc[500]{Human-centered computing~User models}
\ccsdesc[300]{Human-centered computing~User studies}
\ccsdesc[300]{Applied computing~Psychology}
\ccsdesc[300]{Computing methodologies~Robotic planning}

\ccsdesc[500]{Human-centered computing~Interactive systems and tools}
\ccsdesc[500]{Human-centered computing~User studies}
\ccsdesc[500]{Computer systems organization~Robotics}

\keywords{Object Rearrangement, Human Preference, Psychological Constructs, Monte Carlo Tree Search, User Study
}

\received{30 Sep 2025}
\received[accepted]{1 December 2025}
\received[revised]{31 December 2026}

\maketitle

\section{Introduction}
\label{sec:introductiontext}
Object rearrangement, the problem of organizing items within a space to achieve a desired configuration \cite{batra_rearrangement_2020}, is a central challenge for service robots operating in everyday environments. Here, a robot must be capable not only of manipulating objects, but also of deciding \emph{where each object should go} in a way that aligns with a user's organizational preferences. Human organizational preferences are diverse (e.g. one person may want mugs by the kettle, while another may prefer them in a cabinet) and one-size-fits-all ~\cite{kapelyukh_scenescore_2023,newman_degustabot_2024} definitions of what an acceptable arrangement is might fail to account for these differences. For robots to be useful in this context, they must be equipped with object rearrangement models that capture the salient criteria behind these preferences and that can adapt to differences across users and scenes, especially in shared environments.

Prior work on the personalization of object rearrangement has aimed to tailor placements to reflect an individual user’s subjective spatial preferences rather than a universal notion of tidiness~\cite{KapelyukhMyRules}. Abdo et al. \cite{abdo_organizing_2016} predicted user-specific groupings via collaborative filtering, while \cite{KapelyukhMyRules} introduced a framework for learning latent embeddings of tidying style from demonstrations. More recent systems approximate user preferences with zero-shot visual prompting of vision--language models \cite{newman_degustabot_2024}, infer them from prior and current scene context \cite{ramachandruni2025personalizedroboticobjectrearrangement, CONSOR_chernova}, or actively query users when demonstrations are ambiguous \cite{wang2024apricotactivepreferencelearning}. While these methods move beyond a 'one-size-fits-all' approach, they do so by implicitly using latent representations that capture an overall preference signal without revealing the underlying factors that shape it. This makes it difficult to both understand why objects are placed where they are or  tune arrangements according to specific priorities (e.g., convenience over aesthetics) or different scenarios without intensive retraining.

To address these limitations, we propose grounding personalized object rearrangement in interpretable constructs that reflect how people organize their environments, while remaining adaptable to variation across users and contexts. Specifically, we formulate a compact representation of human organizational preferences in terms of four constructs: spatial practicality, habitual convenience, semantic coherence, and commonsense appropriateness, and investigate whether these human-aligned constructs are sufficient to explain how people reason about object arrangements in common household spaces. Our work makes three contributions:
\begin{itemize}
    \item \textbf{Interpretable formulation of arrangement preferences:} We show that the four explicit arrangement constructs (spatial, habitual, semantic, commonsense) capture variation across individuals and scenarios (i.e. kitchen and living room).
    \item \textbf{A measurement tool for the proposed constructs:} We design and validate a self-report questionnaire that quantifies how strongly each construct influences participants’ judgments and establish that the constructs form a reliable and psychologically meaningful basis.
    \item \textbf{Preferences-alinged arrangement generation:} We formulate cost functions for the constructs and integrate these into a Monte Carlo Tree Search (MCTS) planner for arrangement. This approach produces arrangements that align with human preferences when using participant-derived weights.
\end{itemize}

\section{Related Work}
Most robotic object rearrangement systems optimize for a single, universal definition of what constitutes a ``good'' organization. In the indoor household environments, e.g., kitchens and living rooms, organization is primarily defined at the object- and room-levels, which are often described in spatial cognition as \emph{figural} and \emph{vista} spaces~\cite{montello_scale_1993,gotz_unified_2025}. These methods use visuo-semantic priors and commonsense reasoning to move objects to plausible locations~\cite{sarch_tidee_2022,kant_housekeep_2022}, minimize spatial flow fields~\cite{goyal_ifor_2022}, learn arrangement cost functions~\cite{kapelyukh_scenescore_2023}, or leverage 3D mapping and semantic search~\cite{trabucco2022simpleapproachvisualrearrangement}. While effective at achieving tidy configurations, these methods cannot account for diverse user-specific organizational styles. In contrast, our work formulates users' object (re)arrangement preference as a combination of four interpretable constructs, which is flexible and capable of accommodating diverse user preferences.

For personalized rearrangement, Abdo et al.~\cite{abdo_organizing_2016} used collaborative filtering to model co-occurrence patterns of object groupings, but this assumes a fixed organizational schema is given \textit{a priori }and thus captures statistical regularities without explaining the underlying rationale. Other approaches to personalized rearrangement extract latent ``tidying styles" from user-arranged scenes~\cite{KapelyukhMyRules}, use large language models to summarize examples into rules~\cite{wu_tidybot_2023}, infer preferred placements from partial arrangements~\cite{CONSOR_chernova}, employ zero-shot vision–language models~\cite{newman_degustabot_2024}, or actively query users to resolve ambiguities~\cite{wang2024apricotactivepreferencelearning}. These advances enable personalization and achieve good predictive performance, but rely on implicit representations that hide the principles guiding the generated arrangements. 

This lack of interpretability limits practical adoption. Reviews in human–robot interaction (HRI) and explainable robotics~\cite{anjomshoae2019explainable,sakai2021explainableautonomousrobotssurvey} emphasize that users (especially in personal spaces) benefit from explanations that communicate a robot’s goals and reasoning in human terms, rather than abstract model outputs. People prefer robots whose actions are legible and explainable~\cite{dragan2013legibility,chakraborti2017plan}. Both robotics research~\cite{rodríguezlera2024roxiedefiningroboticexplanation} and broader AI contexts~\cite{rudin2019stopexplainingblackbox} increasingly recognize that inherently interpretable models are preferable to black-box systems requiring post-hoc explanation, particularly when trust and transparency affect adoption. We address these drawbacks by explicitly formulating arrangement preferences along four interpretable constructs. This design provides two key benefits. First, it enables transparent characterization of individual and group organizational styles within a unified framework. Second, it provides a foundation for robots that can personalize behavior and communicate reasoning using simple, understandable terms.

\section{Methodology}

To address the lack of interpretable constructs in current robotic object rearrangement research, we propose four constructs motivated by human organizational reasoning as detailed in Sec~\ref{subsec:motivation}. We validate the proposed constructs with a user study detailed in Sec~\ref{sec:userstudy}, and demonstrate how they can be used for computational generation of human-like arrangements as detailed in Sec~\ref{sec:method-formulation}.

\subsection{Theoretical Motivation}
\label{subsec:motivation}

Inspired by the analysis of psychological designs involving spatial cognition, ergonomics, and human–environment interaction, and the reviewing on robotics literature, we propose four constructs to provide comprehensive coverage of human organizational reasoning: \emph{spatial practicality}, \emph{habitual convenience}, \emph{semantic coherence}, and \emph{commonsense appropriateness}. 

\textbf{Spatial practicality} captures how people place items in locations that fit the physical layout of the room and support efficient, physically feasible use of the arranged objects. Because our scenarios involve indoor kitchens and living rooms, we focus on organization at \emph{figural} and \emph{vista} spatial scales, that is, object-to-surface relations and within-room layouts, rather than larger environmental navigation scales~\cite{montello_scale_1993, gotz_unified_2025}. At these indoor scales, research on \emph{scene grammar} shows that people learn regularities about where objects typically appear relative to functional regions and stable anchors (e.g., sinks or stoves)~\cite{vo2019reading}, and that violations of these regularities reduce perceived plausibility and can incur measurable processing costs~\cite{8575436, draschkow_scene_2017,vo_differential_2013}. Contextual cueing studies demonstrate that people implicitly learn recurring spatial configurations and use them to guide expectation and attention during visual search~\cite{chun1998contextual}. In robotics, related ideas appear in object-placement systems that evaluate candidate placements using geometric structure and physical feasibility criteria (e.g., support contact, stability, etc.), including learning-based placing from 3D point clouds and planners that search for stable poses on available surfaces~\cite{jiang_learning_2012,harada2012objectplacement,paxton2022predicting}.

\textbf{Habitual convenience }reflects how people make frequently used items easy to reach. Actions repeated in the same environment become automatic rather than deliberate~\cite{wood2016psychology,NEAL2012492}. Neuroscience research shows that familiar environments trigger these automatic behaviors instead of conscious decision-making~\cite{graybiel2008habits}. This creates a natural drive to minimize effort for routine tasks by positioning frequently used objects within easy reach. This principle is also used in design guidelines and ergonomic standards, which often recommend placing high-use items in primary reach zones to reduce physical strain~\cite{yang2009human,pheasant_bodyspace_2018,kroemer_fitting_2008}. Manufacturing guidelines like 5S apply the same logic, organizing tools by usage frequency to eliminate wasted motion~\cite{hirano_5_1995}. 

\textbf{Semantic coherence} emerges when people place items together if they are used for the same task or are contextually related. People tend to group objects that participate in the same activities because our brains link them through functional relationships~\cite{o2018semantic}. Prior research also shows humans classify environments primarily by the activities they afford rather than how things look~\cite{greene_recognition_2009}. As a result, items used together become mentally chunked as units, improving both memory retrieval and search efficiency~\cite{o2018semantic}. This follows associative learning principles under which items that regularly co-occur in our experience become mentally linked and are treated as belonging together by the brain~\cite{anderson2014human}. Recent approaches to object rearrangement in robotics, such as \textit{ConSOR}~\cite{CONSOR_chernova} and \textit{ContextSortLM}~\cite{ramachandruni2025personalizedroboticobjectrearrangement}, exploit this semantic context by grouping objects according to their functional relationships and organizational schemas.

\textbf{Commonsense appropriateness} drives people to put things where others usually expect to find them. Humans rapidly detect when objects are ``out of place'' because of internalized expectations about what belongs where~\cite{8575436, vo2019reading,doi:10.1177/0956797613476955}. These expectations often reflect accumulated wisdom about widely accepted safety, hygiene, and social norms. This construct is compelling because, while aesthetic preferences might vary across cultures~\cite{CHE201877}, many basic safety and social norms (e.g. placing heavy objects on stable surfaces, placing utensils near where they are used, keeping cleaning chemicals away from food) are more standard ~\cite{fda_foodcode_2022,rch_poisoning_2025,oliva_role_2007}. Systems like \textit{TIDEE}~\cite{sarch_tidee_2022} achieve human-like tidying performance precisely by respecting these fundamental normative constraints, demonstrating that commonsense rules can be learned and applied systematically.

\subsection{User Study}
\label{sec:userstudy}

\begin{figure*}[t]
    \centering
    \includegraphics[width=\textwidth]{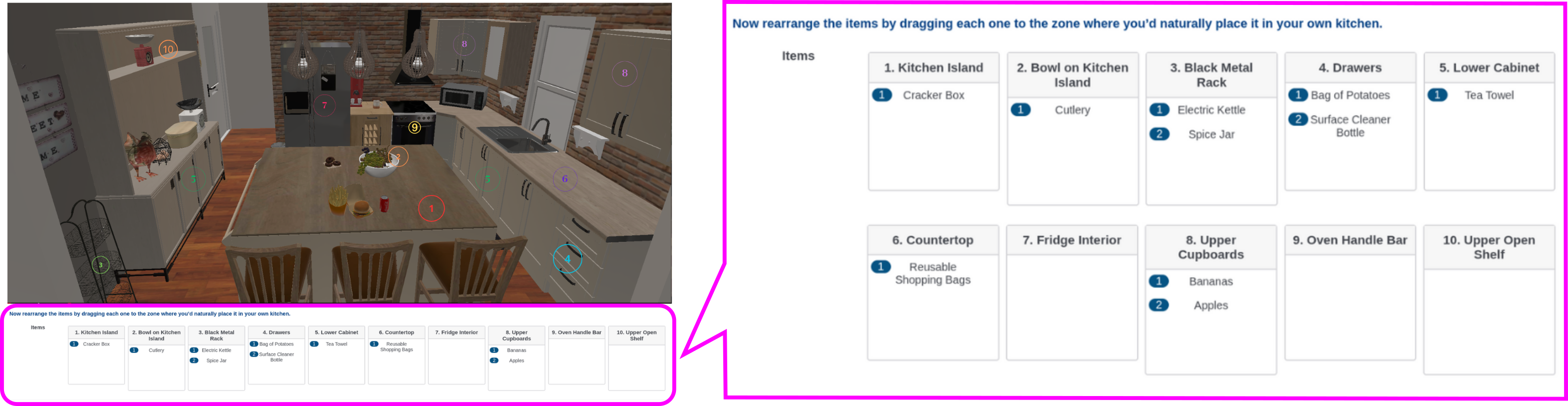}
    
    \Description{Survey interface showing an example organization task in the kitchen. Participants arranged objects by dragging them into available receptacle zones. More details about the interface are provided in the appendix.}
    
    \caption{Survey interface showing an example organization task in the kitchen. Participants arranged objects by dragging them into available receptacle zones. More details about the interface are provided in the appendix.}
    \label{fig:survey_interface}
\end{figure*}
\begin{table}[t]
\centering
\footnotesize
\renewcommand{\arraystretch}{0.92}
\setlength{\tabcolsep}{2.5pt}
\caption{Survey instrument: constructs of organizational preference with extended-form items.}
\label{tab:principle_items}
\begin{tabular}{p{0.18\columnwidth} p{0.78\columnwidth}}
\toprule
\textbf{Construct} & \textbf{Extended Form (3 items)} \\
\midrule
\multirow{3}{=}{Spatial\\Practicality}
& I had a clear spot in mind for each item. \\
& I avoided placements that felt awkward or out of place. \\
& I tried to place items as close as possible to their ideal spot. \\
\midrule
\multirow{3}{=}{Habitual\\Convenience}
& I placed each item based on my everyday routine. \\
& I made sure the items I use most often were easier to grab. \\
& I considered how often I use each item when deciding placement. \\
\midrule
\multirow{3}{=}{Semantic\\Coherence}
& I placed items near each other if they are usually used together. \\
& I placed items together if they served a similar purpose. \\
& I avoided grouping items that do not belong near each other. \\
\midrule
\multirow{3}{=}{Commonsense\\Appropriateness}
& I placed items where most people would expect to find them. \\
& I used what I’ve generally learned about how rooms are organized. \\
& I avoided placements that would look messy or unusual to others. \\
\bottomrule
\end{tabular}
\end{table}

\label{user_study_design}
We conducted an online user study aiming to validate the proposed constructs via Qualtrics\footnote{https://www.qualtrics.com}. We adopted a within-subjects design in which each participant completed four organization tasks: two in the kitchen and two in the living room, (Fig.~\ref{fig:survey_interface}). In each scenario, participants performed Task~1, arranging a set of objects from scratch, and Task~2, re-arranging a pre-existing configuration into a layout they found preferable. Two distinct scenarios were used to determine whether selected constructs generalise across settings, while task variations were selected to increase measurement validity.

Participants interacted with pre-rendered household scenes drawn from the Habitat Synthetic Scenes Dataset (HSSD-200D)~\cite{khanna2023habitatsyntheticscenesdataset}. Each scenario contained a fixed set of objects and receptacles chosen to reflect realistic organization challenges. They were tasked with placing each object into one of the available receptacle zones in the scene, using a drag-and-drop interface, to create an arrangement that felt natural and appropriate (see Fig.~\ref{fig:survey_interface}). After completing each task, participants rated their satisfaction, from 0 to 100, with the resulting arrangement (with both pre- and post-ratings collected in Task~2). 

Measures were collected on a 5-point Likert scale along the four proposed constructs of organizational preference, introduced in Section~\ref{subsec:motivation}. To capture these, participants rated their agreement with three items per construct (12 items total). To minimize potential bias, the constructs themselves were never presented explicitly to participants; instead, items were phrased as natural self-reflection statements (e.g., ``I placed each item based on my everyday routine'').
Table~\ref{tab:principle_items} summarizes the constructs and the corresponding extended-form items. 

In addition to these structured ratings, the survey included several open-ended prompts asking participants about what influenced their satisfaction ratings, what additional factors may have shaped their placement decisions, and whether any aspect of the task felt difficult or unnatural. A final prompt invited participants to share any additional reflections about how they organized items across tasks or about the survey in general. These open-ended prompts allowed participants to articulate considerations beyond the four proposed constructs, ensuring that emergent factors could be captured and qualitatively analyzed. Attention checks were also included to maintain engagement and detect poor quality submissions. 

We recruited a total of $N=63$ participants through the Prolific online crowd-sourcing platform and institutional networks. Participants were required to be at least 18 years old and proficient in English. Participants had a mean age of $M=32$ years ($SD=13$), spanning the 18–65+ range.  Recruitment and study procedures were approved by Monash University Human Research Ethics Committee (ID: 47370) prior to data collection. Participants provided informed consent and received £3 for a median completion time of $\sim$20 minutes.

\subsection{Computational Generation of Human-Aligned Arrangements}
\label{sec:method-formulation}
The four proposed constructs can be formulated as cost functions within a personalised object rearrangement task. We model personalized object rearrangement as the task of assigning a set of objects $\mathcal{O}=\{o_i\}_{i=1}^N$ to a set of receptacles $\mathcal{R}=\{\rho_j\}_{j=1}^M$. 
Arrangements are represented as a set of object–receptacle placements:
\begin{equation}
X = \{ (o_i, \rho_j, v_i) \mid o_i \in \mathcal{O}, \, \rho_j \in \mathcal{R}, \, v_i \in P_j \}, \label{eq:arrangement}
\end{equation}
where $v_i$ is the placement position of object $o_i$ on receptacle $\rho_j$, and $P_j$ is the valid placement surface of $\rho_j$. 
Feasible arrangements $\mathcal{F}$ must satisfy the following: \textbf{unique assignment:} each object is placed exactly once; \textbf{surface containment:} $v_i \in P_j$ for all placements $(o_i,\rho_j,v_i)$; \textbf{non-overlap:} objects placed on the same receptacle do not intersect in 3D space. Arrangement quality is evaluated through four normalized scoring functions $\{f_k(X)\}_{k=1}^4$ with outputs in $[0,1]$ corresponding to the constructs introduced in Sec.~\ref{subsec:motivation}. We mathematically instantiate these constructs as follows:

\textbf{Spatial practicality}, where $v_i$ is the current placement of object $o_i$ and $v_i^\star$ is a preferred prior location inferred from demonstrations:
\begin{equation}
f_1(X) = \frac{1}{N} \sum_{i=1}^{N} \frac{1}{1 + \| v_i - v_i^\star \|}.
\end{equation}

\textbf{Habitual convenience}, where $u_{\max} = \max_{i=1,\dots,N} u_i$ is used to normalize usage frequency, and $\alpha_j \in [0,1]$ denotes the accessibility of receptacle $\rho_j$ ( higher values indicate greater accessibility): 
\begin{equation}
f_2(X) = 1 - \frac{1}{N} \sum_{i=1}^{N} \left( \frac{u_i}{u_{\max}} - \alpha_j \right)^2.
\end{equation}

\textbf{Semantic coherence}, with $d_{ij} = \| v_i - v_j \|$ the distance between objects $o_i$ and $o_j$. Object affinities $\sigma_{ik} \in [-1,1]$ are estimated from demonstrations, usage statistics, or semantic knowledge bases:
\begin{equation}
f_3(X) = 1 - \frac{1}{N(N-1)} \sum_{i=1}^{N} \sum_{\substack{j=1 \\ j \neq i}}^{N}
\begin{cases}
\sigma_{ij} \cdot \frac{d_{ij}}{1 + d_{ij}}, & \sigma_{ij} > 0 \\
|\sigma_{ij}| \cdot \left(1 - \frac{d_{ij}}{1 + d_{ij}} \right), & \sigma_{ij} < 0
\end{cases}.
\end{equation}

\textbf{Commonsense appropriateness} $\in [0,1]$ estimated by querying language model conditioned on the current arrangement state $s_t$ and a JSON description of objects and receptacles:
\begin{equation}
f_4(X) = \frac{1}{N} \sum_{i=1}^{N} \text{commonsense\_score}(o_i, \rho_j; s_t)
\end{equation}.

These functions are aggregated into a scalar reward:
\begin{equation}
R(X; \mathbf{w}^{(k)}) = \sum_{m=1}^4 w^{(k)}_m f_m(X), \label{eq:reward}
\end{equation}

where $\mathbf{w}^{(k)} = \left[ w^{(k)}_1,\dots,w^{(k)}_4 \right] \in [0,1]^4$ denotes the preference vector of user $k$ that captures how strongly a user prioritizes each organizational construct. Given a vector $\mathbf{w}^{(k)}$ encoding a user's organizational preferences, the objective is then to find a sequence of placement actions that produces arrangements reflecting these preferences by maximizing the corresponding weighted reward function in Eq.~\eqref{eq:reward}. Formally, this is formulated as:
\begin{equation}
X^{*(k)} = \arg\max_{X \in \mathcal{F}} R(X; \mathbf{w}^{(k)}). \label{eq:optimization}
\end{equation}
where $\mathcal{F}$ denotes the set of all feasible arrangements. 
\label{sec:mcts}

We employ Monte Carlo Tree Search (MCTS)~\cite{swiechowski_monte_2023} to efficiently explore the combinatorial space of object-to-receptacle assignments. MCTS is well-suited to this domain due to its ability to balance exploration and exploitation in large discrete action spaces, making it an effective method to find high-quality arrangement policies.

We specify a user profile with a ground-truth preference vector $\mathbf{w}_{\text{gt}}^{(k)} = \left[w_1^{(k)}, w_2^{(k)}, \dots,
 w_4^{(k)}\right]$,
where each $w_i^{(k)}$ denotes the importance assigned to the $i$-th organizational principle by user $k$; along with construct-specific priors which are estimated separately and held constant during planning (Sec.~\ref{sec:MCTS_results}). Given this profile, MCTS constructs a search tree where nodes correspond to partial arrangements $X_t$ at time step $t$ and edges represent actions of assigning unplaced objects to valid receptacle locations. At each step $t$, the admissible action space $\mathcal{A}(X_t)$ is state-dependent, consisting of all feasible placements of currently unplaced objects. At time step $t$, the tree policy selects an action $a_t$ using the Upper Confidence Bound (UCB) criterion:
\begin{equation}
a_t = \arg\max_{a \in \mathcal{A}(X_t)} \left(Q(X_t, a) + c\sqrt{\frac{\ln n(X_t)}{n(X_t, a)}}\right),
\end{equation}
where $Q(X_t, a)$ is the empirical action-value estimate computed as the mean return from rollouts initiated with $(X_t,a)$ using Eq.~\ref{eq:optimization}, $n(X_t)$ and $n(X_t,a)$ are the visit counts for state  $X_t$ and state–action pair $(X_t, a)$, and $c>0$ controls the exploration–exploitation trade-off. We set $c = 1/\sqrt{2}$, following the UCT analysis in~\cite{10.1007/11871842_29}, which establishes this value under rewards bounded in $[0,1]$. The action selection process terminates once every object has been placed. We set horizon length to $T=N$, i.e., the number of objects, to ensure that different action sequences leading to the same final configuration are equivalent and prevent degenerate behaviors such as reward-hacking through repeated placements.

The best action, i.e., assigning an object to  a valid receptacle, at each node, i.e., object, is determined using
\begin{equation}
a^{*}_{t}(X_t) = \arg\max_{a \in \mathcal{A}(X_t)} \frac{\text{TotalReward}(X_t,a)}{n(X_t,a)},
\end{equation}
where $\text{TotalReward}(X_t,a)=\sum_{k=1}^{n(X_t,a)}\sum_{t}^{T}R(X_t;w^{(k)})$ is the accumulated reward over rollouts starting from state-action pair $(X_t,a)$.

The resulting sequence of actions defines a trajectory:
\begin{equation}
\pi^{*} = [a^{*}_{1}(X_1), a^{*}_{2}(X_2), \dots, a^{*}_{T}(X_{T})],
\end{equation}

that showcases how arrangements consistent with a given preference profile can be realized.

\section{Results and Discussion}
\label{sec:results}
We structure our results to answer complementary questions about the proposed constructs and their roles in explaining arrangement preferences. First, Sec.~\ref{EFA_val} validates the questionnaire and examines whether participants’ responses organize into coherent factors aligned with the four constructs. Next, Sec.~\ref{sec:formulation_validation} tests whether variation in these construct ratings is reflected in participants’ reported satisfaction with arrangements, establishing their explanatory value. Finally, Sec.~\ref{sec:MCTS_results} shows how participant-derived construct weights can be operationalized within our planning framework to generate arrangements that better align with human preferences. All statistical analyses (factor analyses, regressions, and nonparametric tests) were conducted in \textit{jamovi} (Version 2.7)~\cite{jamovi2025} and R (Version 4.5)~\cite{Rcore2025}, with regression models estimated using GAMLj~\cite{gallucci2019gamlj}.

\subsection{Psychometric Validation of Questionnaire}
\label{EFA_val}
To assess whether the questionnaire provides a reliable basis for measuring the four proposed constructs, we conducted an Exploratory Factor Analysis (EFA)~\cite{fabrigar1999evaluating} on the 12 Likert items (three per construct) in Table~\ref{tab:principle_items}. Responses from all task–scene combinations were included, yielding four observations per participant. The analysis used minimum residual extraction with oblimin rotation~\cite{costello2005best}, which is recommended when psychological constructs are expected to correlate rather than be strictly orthogonal. This choice was also consistent with the observed inter-factor correlations in our data, which fell in the moderate range ($r = 0.30 \text{ to } 0.48$). Data suitability checks were conducted before EFA, where Bartlett’s test of sphericity ($\chi^2(66) = 660$, $p < 0.001$) indicates the variables are significantly correlated and suitable for EFA, and the  Kaiser–Meyer–Olkin (KMO) measure of sampling adequacy with a value 0.80 (0.5 is considered a minimum acceptable threshold) also supported the use of EFA.

Factor retention was guided by parallel analysis and the inspection of the scree plot, both supporting a four-factor solution~\cite{hayton2004factor,ledesma2007determining}.  Together, these factors explained 61\% of the variance in participants’ item responses. The factor loadings broadly aligned with the hypothesized structure: semantic items loaded strongly on a single factor ($0.70$–$0.92$), spatial items clustered together ($0.55$–$0.78$), and habitual ($0.62$–$0.81$) and commonsense ($0.34$–$0.93$) items generally grouped as expected, though with greater variability. Higher loading values indicate a stronger correspondence between an item and its intended underlying construct and can be taken as evidence that the questionnaire items functioned as intended. 

EFA analysis also revealed two instances in which one item reported  significant factor loadings on more than one construct. First, one habitual item loaded on the spatial factor ($0.62$). Given the substantial inter-factor correlation between spatial and habitual factors ($r = 0.40$), this pattern is interpretable as conceptual overlap, that is, routine-driven placement decisions often involve some notion of spatial reasoning (e.g., people may habitually store coffee mugs near the kettle, a choice that is both convenient for daily use and spatially logical relative to the appliance). Second, one commonsense item showed near-equal loadings on both spatial ($0.35$) and commonsense ($0.34$) factors. This cross-loading may reflect ambiguity in item wording but is also consistent with the observed correlation between spatial and commonsense factors ($r = 0.40$), suggesting these constructs are likely related.  Internal consistency was acceptable to excellent across the four scales (Cronbach’s $\alpha$~\cite{taber2018use}: Spatial = 0.72, Habitual = 0.73, Semantic = 0.87, Commonsense = 0.75), indicating that items within each scale exhibited correlated response patterns that reliably measured the same underlying construct. 

Overall, the EFA successfully validated the four constructs introduced in Sec.~\ref{sec:method-formulation}. Items designed for each construct clustered together as hypothesized, and they  explained a substantial proportion of ratings variance. However, observed cross-loadings suggest that participants' reasoning about organizational constructs is intertwined. For example, storing mugs near a kettle reflects both habitual convenience and spatial logic, while placing heavy items low involves both commonsense safety and spatial coherence. 

\subsection{Empirical Validation of Proposed Constructs} 
\label{sec:formulation_validation}
We assumed that organizational preferences can be represented by four shared constructs and that individuals differ in how they prioritize them. To evaluate this claim, we tested (i) whether variation along these constructs is associated with participants’ reasoning about arrangements, and (ii) whether participants’ arrangements exhibit heterogeneity consistent with individualized preferences.

\begin{table*}[htbp]
\centering
\caption{GLMMs results predicting satisfaction from each factor. Odds ratios (OR) and 95\% confidence intervals (CI) reported. Model fit: marginal $R^2$ reflects variance explained by fixed effects; conditional $R^2$ reflects variance explained by the full model including random effects; ICC quantifies participant-level heterogeneity. All predictors $z$-scored.}
\label{tab:regression_results}
\begin{tabular}{lcccccccc}
\toprule
\textbf{Factor} & $\hat{\beta}$ & SE & OR [95\% CI] & $z$ & $p$ & Marginal $R^2$ & Conditional $R^2$ & ICC \\
\midrule
Spatial     & 1.10 & 0.30 & 3.01 [1.67, 5.44] & 3.65 & \textbf{$<$0.001} & 0.20 & 0.73 & 0.66 \\
Habitual    & 0.64 & 0.25 & 1.90 [1.17, 3.09] & 2.59 & \textbf{0.010}    & 0.12 & 0.70 & 0.66 \\
Semantic    & 1.13 & 0.37 & 3.10 [1.49, 6.45] & 3.03 & \textbf{0.002}    & 0.19 & 0.77 & 0.71 \\
Commonsense & 0.48 & 0.30 & 1.62 [0.91, 2.90] & 1.64 & 0.102    & 0.10 & 0.71 & 0.67 \\
\bottomrule
\end{tabular}
\end{table*}

\begin{table*}[t]
\centering
\scriptsize
\renewcommand{\arraystretch}{0.95}
\setlength{\tabcolsep}{3pt}
\caption{Qualitative analysis: merged open codes and illustrative participant quotes.
S = satisfaction response; P = placement response.
Full codebook and extended quotes are available in Appendix~A.}
\label{tab:qual_examples_combined}
\begin{tabularx}{\textwidth}{@{}p{0.09\textwidth} >{\RaggedRight\arraybackslash}p{0.30\textwidth} X@{}}
\toprule
\textbf{Construct} &
\textbf{Example open codes (satisfaction / placement)} &
\textbf{Representative quotes} \\
\midrule
Spatial &
access\_reach, workflow\_proximity, save\_space, design\_affordance /
access\_reach, design\_affordance, exit\_location, proximity\_task &
\textbf{(S)} “Mix of practicality, safety, and how often I use them.” \newline
\textbf{(P)} “I would place items where they would be easiest to reach and most convenient for the task.” \\[3pt]
\midrule
Habitual &
habit\_schema, freq\_use, label\_confusion /
habit\_schema, freq\_use, less\_used\_far, memory\_findability &
\textbf{(S)} “Initial arrangement non-functional … reorganized based on how I function in my kitchen.” \newline
\textbf{(P)} “The less often used items can be stored further away … the controller can be stored on the fireplace mantle.” \\[3pt]
\midrule
Semantic &
semantic\_grouping, context\_unknown, clutter\_risk /
semantic\_grouping, canonical &
\textbf{(S)} “A few items don't fit this room overall.” \newline
\textbf{(P)} “Things like spoons in a block naturally go on the top … useful to have basics out to grab while cooking.” \\[3pt]
\midrule
Commonsense &
perishability, hygiene\_safety, workflow\_proximity /
perishability, hygiene\_safety, safety &
\textbf{(S)} “Worked well — keeping the kitchen island clear … toddler … prefer drawers.” \newline
\textbf{(P)} “I also make sure that perishable items are either in the fridge or in a cupboard (not on the counter where they'll spoil).” \\[3pt]
\midrule
Emergent &
aesthetics, clutter\_risk, personal\_constraint, label\_confusion /
aesthetics, social\_others, temporary\_use, personal\_constraint &
\textbf{(S)} “Spice jars away from stove didn't work; prefer drawer for ease (height).” \newline
\textbf{(P)} “Because I saw many of the items as ones I would not store in the living room … I placed them on the coffee table.” \\
\bottomrule
\end{tabularx}
\end{table*}

\subsubsection{Shared Dimensions as a Basis for Assessing Arrangements}\label{subsec:results_rq2}

We perform a linear regression analysis to examine whether the proposed four constructs had an effect on participants’ satisfaction ratings for each arrangement task. Specifically, we test whether variations in these dimensions' ratings are statistically associated with differences in reported satisfaction. We also analyzed participants' responses to open-ended questions to check alignment with participants’ stated reasoning and identify any considerations outside our hypothesized set.

\textbf{Regression Analysis:} We fit Generalized Linear Mixed Models (GLMMs) for ordinal outcomes, using proportional odds models with satisfaction ratings recoded into three categories (low, medium and high) via quantile binning at the 33rd and 66th percentiles to mitigate skew in the continuous scale. The aim of this step was not to claim that ``more of a given principle always yields higher satisfaction’’ in any universal sense, but rather to assess whether variation along these dimensions was statistically associated with satisfaction. 

For each latent construct
$$Z \in \{\text{Spatial, Habitual, Semantic, Commonsense}\},$$ we estimated a proportional-odds mixed model of the form:
\begin{equation}
\begin{split}
\text{logit}\!\left(P(Y_{ij} \leq k)\right) 
&= \theta_k 
 - \big( \beta_1 Z_{ij} + \beta_2 \,\text{room/task}_{ij} \\
& \quad + \beta_3 \,(Z_{ij} \times \text{room/task}_{ij}) + u_i \big),
\end{split}
\label{eq:glmm_model}
\end{equation}

where $Y_{ij}$ is the satisfaction rating for participant $i$ on observation $j$, $k \in \{1,2\}$ indexes the ordinal thresholds, with $\theta_1$ separating Low from Medium/High and $\theta_2$ separating Low/Medium from High, $Z_{ij}$ is the construct predictor, \textit{room/task}$_{ij}$ is a fixed effect for scene $\times$ task, and $u_i$ is a random intercept for participant ($u_i \sim \mathcal{N}(0,\sigma^2)$).

Given the partial cross-loadings and moderate correlations observed in the EFA (Sec. ~\ref{EFA_val}), we fit separate models for each construct to ensure clearer interpretation. Results (Table~\ref{tab:regression_results}) show that three of the four hypothesized constructs were significantly associated with satisfaction. Spatial ratings were the strongest predictor ($\hat{\beta}=1.10$, OR = 3.01, 95\% CI [1.67, 5.44], $p<0.001$), indicating that a 1-standard-deviation increase in spatial alignment was associated with roughly tripled odds of reporting higher satisfaction. Habitual ratings were also predictive ($\hat{\beta}=0.64$, OR = 1.90, 95\% CI [1.17, 3.09], $p=0.010$), though with smaller effect size. Semantic ratings had a similar effect magnitude to Spatial ($\hat{\beta}=1.13$, OR = 3.10, 95\% CI [1.49, 6.45], $p=0.002$). All models converged successfully with an acceptable fit. Marginal $R^2$ values (0.10–0.20) indicated that fixed effects explained modest variance, while high conditional $R^2$ values (0.70–0.77) and ICCs (0.66–0.71) allude to substantial variation among participants' satisfaction ratings. Commonsense ratings showed a positive trend but were not statistically significant as a unique predictor ($\hat{\beta}=0.48$, OR$=1.62$, 95\% CI $[0.91, 2.90]$, $p=0.102$). Given the moderate inter-factor correlations and cross-loading patterns observed in the EFA (Sec.~\ref{EFA_val}), this result is consistent with the fact that commonsense appropriateness is often applied \emph{alongside} other reasoning modes in everyday organization. Normative judgments about what is safe, hygienic, or socially appropriate frequently co-occur with spatial feasibility (e.g., reachable and stable placements), habitual accessibility, or semantic grouping, so their explanatory variance is shared. As a consequence, the GLMM coefficient for commonsense can be attenuated even when commonsense reasoning is active.

Overall, these results suggest that satisfaction judgments varied systematically with spatial, habitual, and semantic principles, while commonsense expectations played a more context-dependent role and were less influential as independent predictors. These differences in predictive strength and substantial participant-level variance support modeling preferences as personalized weightings over a shared basis of latent constructs.

\textbf{Qualitative Analysis of Reasoning}
We analyzed the free-text responses from participants who provided reasoning for their arrangement decisions. Participants explained both their satisfaction ratings and placement considerations across room--task contexts, yielding 118 satisfaction reasoning responses and 47 placement consideration responses. We employed a two-stage inductive--deductive coding procedure to analyze participants' responses. First, responses were coded openly using thematic analysis~\cite{Braun01012006}. Second, codes were mapped to our four constructs, with unmapped codes retained as emergent categories. This approach allowed us to confirm whether the hypothesized constructs spontaneously emerged in participants’ reasoning, as well as identify additional themes as potential extensions for future modeling. Table~\ref{tab:qual_examples_combined}shows this mapping with open codes and illustrative participant quotes.

\emph{Spatial} considerations dominated both satisfaction reasoning (57\%) and placement considerations (60\%). \emph{Habitual} factors appeared consistently (31\% and 30\% respectively), while \emph{Semantic} reasoning was more prominent in satisfaction judgments (27\%) than placement decisions (4\%). \emph{Commonsense} appeared infrequently (7\% and 15\%), typically combined with other principles rather than independently. These patterns mirror our quantitative findings where spatial and habitual were strongest predictors, semantic played a secondary role, and commonsense showed context-dependent effects. Constructs frequently co-occurred rather than appearing in isolation. For instance, spatial reasoning commonly paired with habitual (17 satisfaction; 10 placement responses) and semantic considerations (15 satisfaction responses). Commonsense rarely appeared independently, instead coupling with spatial or habitual factors. This qualitative pattern is consistent with the ``filtering'' role for commonsense: normative constraints (e.g., safety or social norms) can rule out otherwise plausible placements, while the remaining variation in satisfaction is more strongly explained by spatial, habitual, and semantic considerations.

Emergent themes appeared in 21--32\% of responses, including \emph{design affordances} (missing hooks, outlets), \emph{label confusion}, \emph{context uncertainty}, \emph{aesthetics}, \emph{social influences}, and \emph{personal constraints}. Many emergent themes represent refinements of core constructs. For instance, design affordances and context uncertainty relate to spatial practicality, while personal constraints like reachability align with habitual convenience. However, themes like aesthetics and social influences extend beyond our framework, suggesting other potential constructs to explore in future work.

Overall, the qualitative analysis strongly supports our hypothesized constructs. Spatial and habitual reasoning dominated participants' explanations, semantic coherence appeared as a consistent secondary factor, and commonsense contributed primarily through combinations with other principles. The four constructs provide a parsimonious foundation for modeling arrangement preferences, capturing stable organizational logic. Emergent themes highlight situational variations that could inform future extensions.

\subsubsection{Behavioral Heterogeneity in Organization}
\label{subsec:behavioral_heterogeneity}

We further hypothesize that participants make individualized placement decisions, reflecting distinct trade-offs in how different organizational considerations are prioritized.
\label{hypo:H1}
To test this, we analyzed (i) the similarity of participants’ final layouts and (ii) the relative importance they assigned to the four hypothesized principles. For each scenario $c = \left\{ (a, b) |\;a\in \left\{ \text{Kitchen}, \text{Living}\right\} \text{and}\; b\in \left\{ \text{Task 1}, \text{Task 2}\right\} \right\}$ (see Sec.~\ref{sec:userstudy}), we represent a participant’s arrangement as a set of object--receptacle assignments $S_p = \{(o,\rho)\}$, which is a simplified version of Eq. \ref{eq:arrangement}, where $o$ denotes an object, $\rho$ is a receptacle, and $p$ indexes a participant. Similarity between any two arrangements $S_p$ and $S_q$ was quantified using the Jaccard similarity index \cite{KOTU201965}.

For each scenario, we computed pairwise Jaccard similarities across participants’ arrangements and reported the mean values with bootstrapped 95\% confidence intervals.
Similarity was consistently low overall ($M = 0.27$, 95\% CI [0.26, 0.28]), indicating substantial variation in how participants organized the same objects. Kitchen scenarios showed modestly higher similarity ($M = 0.33$, 95\% CI [0.32, 0.34]) compared to living room scenarios ($M = 0.22$, 95\% CI [0.21, 0.23]). This difference likely reflects stronger functional constraints in kitchens, where established conventions dictate logical placements, e.g., placing cooking utensils near the stove or storing dishes near the sink. Living rooms, by contrast, offer greater flexibility in object arrangement, as items like books, decorations, or electronics can be placed in multiple locations without violating clear functional principles. Task type had minimal impact: whether participants arranged objects from scratch (Task~1) or modified an existing layout (Task~2) yielded similar agreement levels within each scenario.

\begin{figure}[t]
  \centering
  \includegraphics[width=0.92\linewidth]{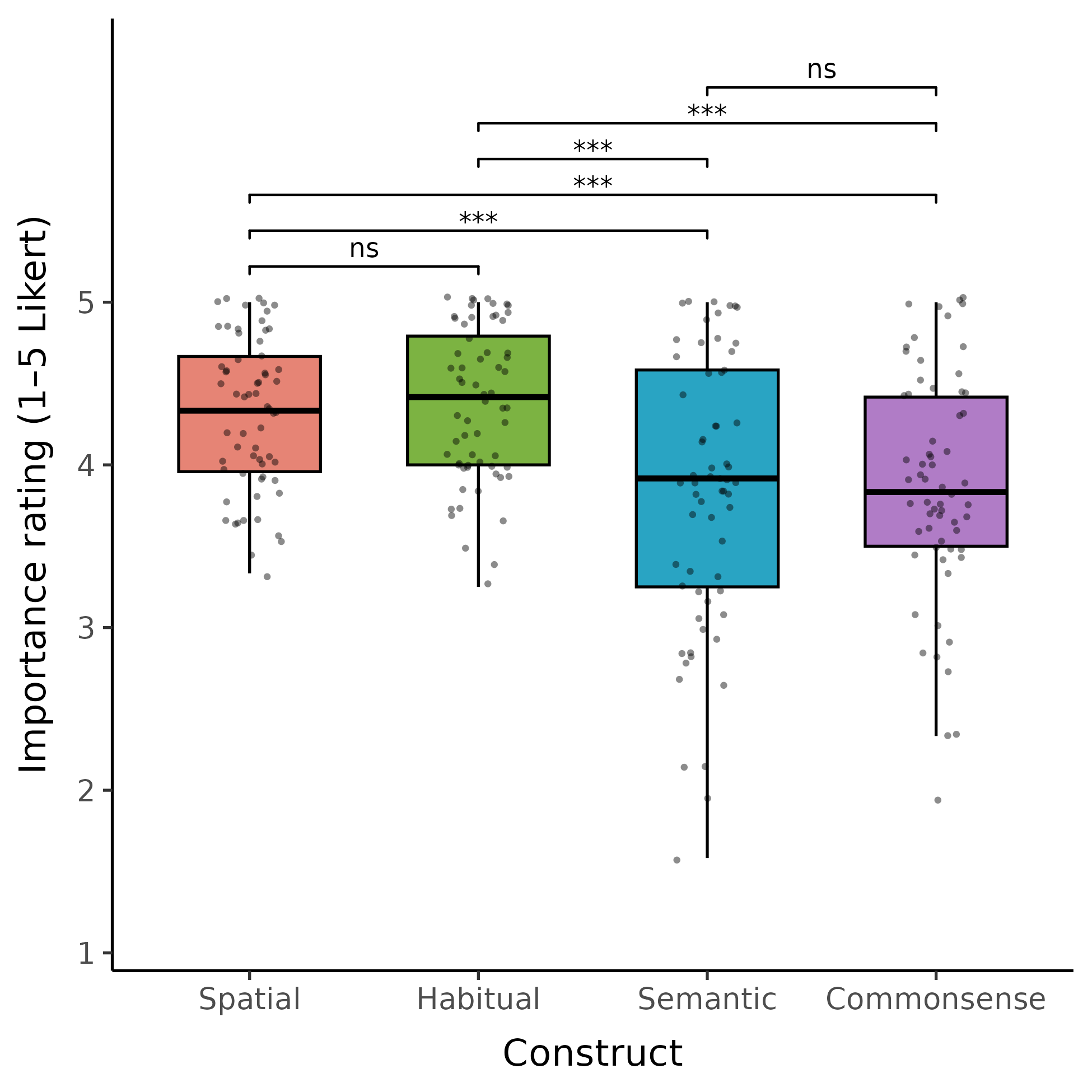}
  \Description{Boxplot of importance ratings (1-5) for four constructs. Spatial and Habitual share high medians (approx. 4.5) with no significant difference between them (ns). Semantic and Commonsense share lower medians (approx. 4.0) and are also not significantly different (ns). However, the high-rated pair (Spatial/Habitual) is significantly higher (p<.001) than the low-rated pair (Semantic/Commonsense).}
  \caption{Importance ratings by construct (boxplots with jitter). Brackets indicate post-hoc comparisons and stars denote $p$ levels (***~$<.001$, **~$<.01$, *~$<.05$, n.s.~$\ge .05$).}
  \label{fig:principle_ratings}
\end{figure}

We further examined whether participants differed in the importance they assigned to the four constructs using a repeated-measures Friedman test on participants’ average construct ratings. The Friedman test revealed significant overall differences in ratings ($\chi^2(3)=154$, $p<0.001$). Post-hoc Durbin--Conover comparisons (Fig.~\ref{fig:principle_ratings}) indicated that \emph{Spatial} and \emph{Habitual} were both rated significantly higher than \emph{Semantic} and \emph{Commonsense} ($p<0.001$ in all cases). Spatial and Habitual also differed slightly ($p{=}0.015$), while Semantic and Commonsense did not ($p{=}0.876$). Both analyses point to strong behavioral heterogeneity. Participant placements showed little similarity, and their construct ratings revealed distinct trade-off patterns: Spatial and Habitual were prioritized, while Semantic and Commonsense were treated as secondary. These results confirm that organizational choices are individual rather than following a fixed canonical template, and that modeling must accommodate user-specific priorities over different arrangement principles.

\subsection{Preference-Aligned Trajectory Generation}\label{sec:MCTS_results}
Given a participant's preference vector $w^{(k)}$ (see Eq.~2), we generated object arrangements by optimizing the weighted sum of four construct-specific scores via MCTS planning (see Sec.~\ref{sec:mcts}). To ensure a fair evaluation, all construct parameters and priors were estimated exclusively from Task~1 participant data. Specifically, spatial priors, receptacle accessibility scores, usage frequencies, and object correlations were estimated from Task~1 placements. Commonsense object--receptacle priors were obtained by querying a large language model (GPT-4) with a structured textual representation (JSON graph) of the scene context and candidate object--receptacle pairs. Preference weights were extracted from questionnaire responses by mapping Likert ratings to numerical scores, averaging items within each construct, and normalizing to yield a personalized weight vector. Generated arrangements were compared against Task~2 participant placements using \textit{object accuracy}, defined as the proportion of objects placed in the same receptacle as a participant.

We selected four representative participant profiles spanning distinct weighting patterns, i.e., spatial-dominant, balanced, habitual-dominant, and semantic-dominant, across the two scenes to evaluate the planner. Table~\ref{tab:profiles} summarizes key characteristics of these profiles, including the scenes, weights, and resulting accuracies ((see Appendix~B for per-object placement details). 

\begin{table}[t]

\centering
\caption{Representative participant profiles used for trajectory generation. 
Weights are normalized across spatial (S), habitual (H), semantic (Se), and commonsense (C).}
\label{tab:profiles}
\begin{tabular}{cp{0.3cm}ccc}
\toprule
\textbf{Pattern} & \textbf{ID} & \textbf{Scene} & \textbf{\makecell{Weight Profile \\(S,H,Se,C)}} & \textbf{\makecell{Accuracy}} \\
\midrule
Spa.-dom.  & P23 & Living R. & [0.37, 0.29, 0.17, 0.17] & 0.60 \\
Balanced     & P32 & Living R. & [0.25, 0.25, 0.25, 0.25] & 0.40 \\
Hab.-dom. & P24 & Kitchen     & [0.34, 0.40, 0.18, 0.08] & 0.80 \\
Sem.-dom. & P16 & Kitchen     & [0.26, 0.21, 0.30, 0.23] & 0.90 \\
\bottomrule
\end{tabular}

\end{table}

\textbf{Spatial-dominant (Living Room, P23).} 
With spatial practicality carrying the most weight, the planner emphasized placing items where they ``naturally fit'' into the room’s layout. This produced successes that aligned with furniture affordances. At the same time, the weaker semantic and commonsense weights meant the system tolerated oddities. These are not random errors but the trade-off of prioritizing layout coherence above category or normative consistency. The overall signature is a room that looks spatially coherent, but with some object groupings that are counterintuitive.  

\textbf{Habitual-dominant (Kitchen, P24).}  
Strong habitual weighting drove the planner to place frequently used objects into the most accessible receptacles. This approach yielded high alignment with participants' placements. Discrepancies arose primarily when the participant's storage preferences diverged from the broader usage patterns. The low commonsense weight prevented the planner from correcting toward more typical placements, while the habitual component, which models receptacle accessibility based on usage frequency, could not account for these associations. This shows how habitual bias produces strong routine fidelity but also exposes the limits of our current modeling when participants deviate from normative patterns.  

\textbf{Semantic-dominant (Kitchen, P16).}  
Semantic grouping dominated this profile, with commonsense moderately supporting it. The planner produced category-faithful groupings that align closely with human expectations and Task~2 ground truth. Errors arose when semantic links to storage locations were weak and commonsense weighting failed to compensate, leading to atypical placements a human would likely avoid. The high accuracy (0.90) demonstrates that semantic bias reliably produces human-like groupings, but is dependent on the completeness of semantic priors.  

\textbf{Balanced (Living Room, P32).} 
This participant reported nearly uniform weights, leaving the planner without a dominant construct to guide the search. While some placements were still correct, other items drifted. Because no construct provided strong direction, the planner explored many near-equal options, producing an arrangement that was acceptable but not tightly structured. The lower accuracy (0.40) may reflect limited guidance from a flat profile \emph{and} the use of estimated hyperparameters and coarse cost-term definitions, which together can mask subtle behavioral biases.  

These examples show that the generated arrangements were reasonably accurate overall, with accuracies ranging from 0.40 to 0.90 depending on the participant’s weight profile (mean $\approx 0.68\% \pm 0.20$). Notably, we observed that the kitchen scene yielded higher accuracies than the living room. This likely reflects that kitchens contain stronger habitual and semantic regularities (e.g., food in fridges, cutlery in drawers) that our cost terms captured well. By contrast, living rooms involve more ambiguous placements where multiple surfaces are equally plausible (e.g., a magazine could be on a coffee table, side table, or shelf), making subtle individual biases harder to model with hyperparameters estimated from limited data. 

\section{Conclusions and Future Work}
We proposed four interpretable constructs of human organizational preference ( i.e., spatial practicality, habitual convenience, semantic coherence, and commonsense appropriateness) and validated them with a user study of 63 participants. Our analyses confirmed that these constructs capture meaningful variation in arrangement preferences across both users and scene contexts (kitchen and living room). Qualitative responses also showed that participants naturally reasoned about their organizational choices in terms consistent with these constructs. We mathematically formulated the constructs as cost functions and integrated them into an MCTS planner guided by participant-specific weight profiles. The generated arrangements mostly aligned with human placements in both quantitative accuracy and qualitative signature, demonstrating that the constructs can be operationalized for planning.

While demonstrating strong alignment with human reasoning, operationalizing the proposed constructs within a computational model required certain simplifying assumptions. Cost function hyperparameters were estimated by proxy rather than learned from demonstrations. Additionally, we treated the constructs as an independent linear combination, despite moderate correlations in the user data suggesting these dimensions can be entangled in human reasoning. While the four constructs accounted for substantial variance in organizational preferences, the residual variation indicates that additional factors may be needed to capture remaining individual idiosyncrasies.

Future work will learn construct parameters and preference weights directly from human demonstrations, refine the formulation to account for construct interactions, and integrate emergent factors from our qualitative analysis, such as aesthetics and design affordances. We also aim to extend the method to handle continuous receptacle configurations. Overall, our results demonstrate that organizational principles can be explicitly modeled and leveraged as interpretable building blocks for personalized robot assistance in household object rearrangement.
\balance
\bibliographystyle{ACM-Reference-Format}
\bibliography{references}


\begin{thebibliography}{57}


\ifx \showCODEN    \undefined \def \showCODEN     #1{\unskip}     \fi
\ifx \showISBNx    \undefined \def \showISBNx     #1{\unskip}     \fi
\ifx \showISBNxiii \undefined \def \showISBNxiii  #1{\unskip}     \fi
\ifx \showISSN     \undefined \def \showISSN      #1{\unskip}     \fi
\ifx \showLCCN     \undefined \def \showLCCN      #1{\unskip}     \fi
\ifx \shownote     \undefined \def \shownote      #1{#1}          \fi
\ifx \showarticletitle \undefined \def \showarticletitle #1{#1}   \fi
\ifx \showURL      \undefined \def \showURL       {\relax}        \fi
\providecommand\bibfield[2]{#2}
\providecommand\bibinfo[2]{#2}
\providecommand\natexlab[1]{#1}
\providecommand\showeprint[2][]{arXiv:#2}

\bibitem[Abdo et~al\mbox{.}(2016)]%
        {abdo_organizing_2016}
\bibfield{author}{\bibinfo{person}{Nichola Abdo}, \bibinfo{person}{Cyrill Stachniss}, \bibinfo{person}{Luciano Spinello}, {and} \bibinfo{person}{Wolfram Burgard}.} \bibinfo{year}{2016}\natexlab{}.
\newblock \showarticletitle{Organizing objects by predicting user preferences through collaborative filtering}.
\newblock \bibinfo{journal}{\emph{The International Journal of Robotics Research}} \bibinfo{volume}{35}, \bibinfo{number}{13} (\bibinfo{date}{July} \bibinfo{year}{2016}), \bibinfo{pages}{1587--1608}.
\newblock
\href{https://doi.org/10.1177/0278364916649248}{doi:\nolinkurl{10.1177/0278364916649248}}
\newblock
\shownote{Publisher: SAGE PublicationsSage UK: London, England}.


\bibitem[Anderson and Bower(2014)]%
        {anderson2014human}
\bibfield{author}{\bibinfo{person}{John~R. Anderson} {and} \bibinfo{person}{Gordon~H. Bower}.} \bibinfo{year}{2014}\natexlab{}.
\newblock \bibinfo{booktitle}{\emph{Human Associative Memory} (\bibinfo{edition}{1st edition} ed.)}.
\newblock \bibinfo{publisher}{Psychology Press}, \bibinfo{address}{New York, NY, USA}. 538 pages.
\newblock
\showISBNx{9781315802886}
\href{https://doi.org/10.4324/9781315802886}{doi:\nolinkurl{10.4324/9781315802886}}


\bibitem[Anjomshoae et~al\mbox{.}(2019)]%
        {anjomshoae2019explainable}
\bibfield{author}{\bibinfo{person}{Sule Anjomshoae}, \bibinfo{person}{Amro Najjar}, \bibinfo{person}{Davide Calvaresi}, {and} \bibinfo{person}{Kary Fr\"{a}mling}.} \bibinfo{year}{2019}\natexlab{}.
\newblock \showarticletitle{Explainable Agents and Robots: Results from a Systematic Literature Review}. In \bibinfo{booktitle}{\emph{Proceedings of the 18th International Conference on Autonomous Agents and MultiAgent Systems}} (Montreal QC, Canada) \emph{(\bibinfo{series}{AAMAS '19})}. \bibinfo{publisher}{International Foundation for Autonomous Agents and Multiagent Systems}, \bibinfo{address}{Richland, SC}, \bibinfo{pages}{1078–1088}.
\newblock
\showISBNx{9781450363099}


\bibitem[Batra et~al\mbox{.}(2020)]%
        {batra_rearrangement_2020}
\bibfield{author}{\bibinfo{person}{Dhruv Batra}, \bibinfo{person}{Angel~X. Chang}, \bibinfo{person}{Sonia Chernova}, \bibinfo{person}{Andrew~J. Davison}, \bibinfo{person}{Jia Deng}, \bibinfo{person}{Vladlen Koltun}, \bibinfo{person}{Sergey Levine}, \bibinfo{person}{Jitendra Malik}, \bibinfo{person}{Igor Mordatch}, \bibinfo{person}{Roozbeh Mottaghi}, \bibinfo{person}{Manolis Savva}, {and} \bibinfo{person}{Hao Su}.} \bibinfo{year}{2020}\natexlab{}.
\newblock \bibinfo{title}{Rearrangement: {A} {Challenge} for {Embodied} {AI}}.
\newblock
\urldef\tempurl%
\url{http://arxiv.org/abs/2011.01975}
\showURL{%
\tempurl}
\newblock
\shownote{arXiv:2011.01975}.


\bibitem[Bayat et~al\mbox{.}(2018)]%
        {8575436}
\bibfield{author}{\bibinfo{person}{Akram Bayat}, \bibinfo{person}{Anubhaw~Kumar Nand}, \bibinfo{person}{Do~Hyong Koh}, \bibinfo{person}{Marta Pereira}, {and} \bibinfo{person}{Marc Pomplun}.} \bibinfo{year}{2018}\natexlab{}.
\newblock \showarticletitle{Scene Grammar in Human and Machine Recognition of Objects and Scenes}. In \bibinfo{booktitle}{\emph{2018 IEEE/CVF Conference on Computer Vision and Pattern Recognition Workshops (CVPRW)}}. \bibinfo{publisher}{IEEE}, \bibinfo{address}{Salt Lake City, UT, USA}, \bibinfo{pages}{1992--1999}.
\newblock
\href{https://doi.org/10.1109/CVPRW.2018.00268}{doi:\nolinkurl{10.1109/CVPRW.2018.00268}}


\bibitem[Braun and Clarke(2006)]%
        {Braun01012006}
\bibfield{author}{\bibinfo{person}{Virginia Braun} {and} \bibinfo{person}{Victoria Clarke}.} \bibinfo{year}{2006}\natexlab{}.
\newblock \showarticletitle{Using thematic analysis in psychology}.
\newblock \bibinfo{journal}{\emph{Qualitative Research in Psychology}} \bibinfo{volume}{3}, \bibinfo{number}{2} (\bibinfo{year}{2006}), \bibinfo{pages}{77--101}.
\newblock
\showeprint{https://doi.org/10.1191/1478088706qp063oa}
\href{https://doi.org/10.1191/1478088706qp063oa}{doi:\nolinkurl{10.1191/1478088706qp063oa}}


\bibitem[Chakraborti et~al\mbox{.}(2017)]%
        {chakraborti2017plan}
\bibfield{author}{\bibinfo{person}{Tathagata Chakraborti}, \bibinfo{person}{Sarath Sreedharan}, \bibinfo{person}{Yu Zhang}, {and} \bibinfo{person}{Subbarao Kambhampati}.} \bibinfo{year}{2017}\natexlab{}.
\newblock \showarticletitle{Plan explanations as model reconciliation: moving beyond explanation as soliloquy}. In \bibinfo{booktitle}{\emph{Proceedings of the 26th International Joint Conference on Artificial Intelligence}} (Melbourne, Australia) \emph{(\bibinfo{series}{IJCAI'17})}. \bibinfo{publisher}{AAAI Press}, \bibinfo{address}{Washington, DC, USA}, \bibinfo{pages}{156–163}.
\newblock
\showISBNx{9780999241103}


\bibitem[Che et~al\mbox{.}(2018)]%
        {CHE201877}
\bibfield{author}{\bibinfo{person}{Jiajia Che}, \bibinfo{person}{Xiaolei Sun}, \bibinfo{person}{V{\'i}ctor Gallardo}, {and} \bibinfo{person}{Marcos Nadal}.} \bibinfo{year}{2018}\natexlab{}.
\newblock \showarticletitle{Cross-cultural empirical aesthetics}.
\newblock In \bibinfo{booktitle}{\emph{The Arts and The Brain}}, \bibfield{editor}{\bibinfo{person}{Julia~F. Christensen} {and} \bibinfo{person}{Antoni Gomila}} (Eds.). \bibinfo{series}{Progress in Brain Research}, Vol.~\bibinfo{volume}{237}. \bibinfo{publisher}{Elsevier}, \bibinfo{address}{Amsterdam, Netherlands}, \bibinfo{pages}{77--103}.
\newblock
\showISSN{0079-6123}
\href{https://doi.org/10.1016/bs.pbr.2018.03.002}{doi:\nolinkurl{10.1016/bs.pbr.2018.03.002}}


\bibitem[Chun and Jiang(1998)]%
        {chun1998contextual}
\bibfield{author}{\bibinfo{person}{Marvin~M Chun} {and} \bibinfo{person}{Yuhong Jiang}.} \bibinfo{year}{1998}\natexlab{}.
\newblock \showarticletitle{Contextual cueing: Implicit learning and memory of visual context guides spatial attention}.
\newblock \bibinfo{journal}{\emph{Cognitive psychology}} \bibinfo{volume}{36}, \bibinfo{number}{1} (\bibinfo{year}{1998}), \bibinfo{pages}{28--71}.
\newblock


\bibitem[Costello and Osborne(2005)]%
        {costello2005best}
\bibfield{author}{\bibinfo{person}{Anna~B. Costello} {and} \bibinfo{person}{Jason Osborne}.} \bibinfo{year}{2005}\natexlab{}.
\newblock \showarticletitle{Best practices in exploratory factor analysis: four recommendations for getting the most from your analysis}.
\newblock \bibinfo{journal}{\emph{Practical Assessment, Research, and Evaluation}} \bibinfo{volume}{10}, \bibinfo{number}{1}, Article \bibinfo{articleno}{7} (\bibinfo{year}{2005}), \bibinfo{numpages}{9}~pages.
\newblock
\href{https://doi.org/10.7275/jyj1-4868}{doi:\nolinkurl{10.7275/jyj1-4868}}


\bibitem[Dragan et~al\mbox{.}(2013)]%
        {dragan2013legibility}
\bibfield{author}{\bibinfo{person}{Anca~D. Dragan}, \bibinfo{person}{Kenton C.~T. Lee}, {and} \bibinfo{person}{Siddhartha~S. Srinivasa}.} \bibinfo{year}{2013}\natexlab{}.
\newblock \showarticletitle{Legibility and predictability of robot motion}. In \bibinfo{booktitle}{\emph{Proceedings of the 8th ACM/IEEE International Conference on Human-Robot Interaction}} (Tokyo, Japan) \emph{(\bibinfo{series}{HRI '13})}. \bibinfo{publisher}{IEEE}, \bibinfo{address}{Piscataway, NJ, USA}, \bibinfo{pages}{301--308}.
\newblock
\showISBNx{9781467330558}
\href{https://doi.org/10.1109/HRI.2013.6483603}{doi:\nolinkurl{10.1109/HRI.2013.6483603}}


\bibitem[Draschkow and Võ(2017)]%
        {draschkow_scene_2017}
\bibfield{author}{\bibinfo{person}{Dejan Draschkow} {and} \bibinfo{person}{Melissa L.-H. Võ}.} \bibinfo{year}{2017}\natexlab{}.
\newblock \showarticletitle{Scene grammar shapes the way we interact with objects, strengthens memories, and speeds search}.
\newblock \bibinfo{journal}{\emph{Scientific Reports}} \bibinfo{volume}{7}, \bibinfo{number}{1} (\bibinfo{date}{Nov.} \bibinfo{year}{2017}), \bibinfo{pages}{16471}.
\newblock
\showISSN{2045-2322}
\href{https://doi.org/10.1038/s41598-017-16739-x}{doi:\nolinkurl{10.1038/s41598-017-16739-x}}
\newblock
\shownote{Publisher: Nature Publishing Group}.


\bibitem[Fabrigar et~al\mbox{.}(1999)]%
        {fabrigar1999evaluating}
\bibfield{author}{\bibinfo{person}{Leandre~R Fabrigar}, \bibinfo{person}{Duane~T Wegener}, \bibinfo{person}{Robert~C MacCallum}, {and} \bibinfo{person}{Erin~J Strahan}.} \bibinfo{year}{1999}\natexlab{}.
\newblock \showarticletitle{Evaluating the use of exploratory factor analysis in psychological research.}
\newblock \bibinfo{journal}{\emph{Psychological methods}} \bibinfo{volume}{4}, \bibinfo{number}{3} (\bibinfo{year}{1999}), \bibinfo{pages}{272}.
\newblock


\bibitem[Gallucci(2019)]%
        {gallucci2019gamlj}
\bibfield{author}{\bibinfo{person}{Marcello Gallucci}.} \bibinfo{year}{2019}\natexlab{}.
\newblock \bibinfo{booktitle}{\emph{GAMLj: General, Mixed, and Generalized Linear Models module for jamovi}}.
\newblock
\urldef\tempurl%
\url{https://gamlj.github.io/}
\showURL{%
\tempurl}


\bibitem[Goyal et~al\mbox{.}(2022)]%
        {goyal_ifor_2022}
\bibfield{author}{\bibinfo{person}{Ankit Goyal}, \bibinfo{person}{Arsalan Mousavian}, \bibinfo{person}{Chris Paxton}, \bibinfo{person}{Yu-Wei Chao}, \bibinfo{person}{Brian Okorn}, \bibinfo{person}{Jia Deng}, {and} \bibinfo{person}{Dieter Fox}.} \bibinfo{year}{2022}\natexlab{}.
\newblock \showarticletitle{{IFOR}: {Iterative} {Flow} {Minimization} for {Robotic} {Object} {Rearrangement}}. In \bibinfo{booktitle}{\emph{2022 {IEEE}/{CVF} {Conference} on {Computer} {Vision} and {Pattern} {Recognition} ({CVPR})}}. \bibinfo{publisher}{IEEE}, \bibinfo{address}{New Orleans, LA, USA}, \bibinfo{pages}{14767--14777}.
\newblock
\showISBNx{978-1-66546-946-3}
\href{https://doi.org/10.1109/CVPR52688.2022.01437}{doi:\nolinkurl{10.1109/CVPR52688.2022.01437}}


\bibitem[Graybiel(2008)]%
        {graybiel2008habits}
\bibfield{author}{\bibinfo{person}{Ann~M Graybiel}.} \bibinfo{year}{2008}\natexlab{}.
\newblock \showarticletitle{Habits, rituals, and the evaluative brain}.
\newblock \bibinfo{journal}{\emph{Annu. Rev. Neurosci.}} \bibinfo{volume}{31}, \bibinfo{number}{1} (\bibinfo{year}{2008}), \bibinfo{pages}{359--387}.
\newblock


\bibitem[Greene and Oliva(2009)]%
        {greene_recognition_2009}
\bibfield{author}{\bibinfo{person}{Michelle~R. Greene} {and} \bibinfo{person}{Aude Oliva}.} \bibinfo{year}{2009}\natexlab{}.
\newblock \showarticletitle{Recognition of natural scenes from global properties: {Seeing} the forest without representing the trees}.
\newblock \bibinfo{journal}{\emph{Cognitive psychology}} \bibinfo{volume}{58}, \bibinfo{number}{2} (\bibinfo{date}{March} \bibinfo{year}{2009}), \bibinfo{pages}{137--176}.
\newblock
\showISSN{0010-0285}
\href{https://doi.org/10.1016/j.cogpsych.2008.06.001}{doi:\nolinkurl{10.1016/j.cogpsych.2008.06.001}}


\bibitem[Götz et~al\mbox{.}(2025)]%
        {gotz_unified_2025}
\bibfield{author}{\bibinfo{person}{Friedrich~M. Götz}, \bibinfo{person}{Daniel~R. Montello}, \bibinfo{person}{Michael E.~W. Varnum}, \bibinfo{person}{Davide Luca}, {and} \bibinfo{person}{Douglas~T. Kenrick}.} \bibinfo{year}{2025}\natexlab{}.
\newblock \showarticletitle{A unified framework integrating psychology and geography}.
\newblock \bibinfo{journal}{\emph{Nature Human Behaviour}} \bibinfo{volume}{9}, \bibinfo{number}{9} (\bibinfo{date}{Sept.} \bibinfo{year}{2025}), \bibinfo{pages}{1780--1792}.
\newblock
\showISSN{2397-3374}
\href{https://doi.org/10.1038/s41562-025-02237-y}{doi:\nolinkurl{10.1038/s41562-025-02237-y}}
\newblock
\shownote{Publisher: Nature Publishing Group}.


\bibitem[Harada et~al\mbox{.}(2012)]%
        {harada2012objectplacement}
\bibfield{author}{\bibinfo{person}{K. Harada}, \bibinfo{person}{T. Tsuji}, \bibinfo{person}{K. Nagata}, \bibinfo{person}{N. Yamanobe}, \bibinfo{person}{H. Onda}, \bibinfo{person}{T. Yoshimi}, {and} \bibinfo{person}{Y. Kawai}.} \bibinfo{year}{2012}\natexlab{}.
\newblock \showarticletitle{Object placement planner for robotic pick and place tasks}. In \bibinfo{booktitle}{\emph{2012 IEEE/RSJ International Conference on Intelligent Robots and Systems (IROS)}} (Vilamoura-Algarve, Portugal). \bibinfo{publisher}{IEEE}, \bibinfo{address}{Piscataway, NJ, USA}, \bibinfo{pages}{980--985}.
\newblock
\showISBNx{978-1-4673-1736-8}
\href{https://doi.org/10.1109/IROS.2012.6385800}{doi:\nolinkurl{10.1109/IROS.2012.6385800}}


\bibitem[Hayton et~al\mbox{.}(2004)]%
        {hayton2004factor}
\bibfield{author}{\bibinfo{person}{James~C Hayton}, \bibinfo{person}{David~G Allen}, {and} \bibinfo{person}{Vida Scarpello}.} \bibinfo{year}{2004}\natexlab{}.
\newblock \showarticletitle{Factor retention decisions in exploratory factor analysis: A tutorial on parallel analysis}.
\newblock \bibinfo{journal}{\emph{Organizational research methods}} \bibinfo{volume}{7}, \bibinfo{number}{2} (\bibinfo{year}{2004}), \bibinfo{pages}{191--205}.
\newblock


\bibitem[Hirano(1995)]%
        {hirano_5_1995}
\bibfield{author}{\bibinfo{person}{Hiroyuki Hirano}.} \bibinfo{year}{1995}\natexlab{}.
\newblock \bibinfo{booktitle}{\emph{5 {Pillars} of the {Visual} {Workplace}}}.
\newblock \bibinfo{publisher}{Productivity Press}, \bibinfo{address}{New York}.
\newblock
\showISBNx{978-0-367-80488-6}
\href{https://doi.org/10.4324/9780367804886}{doi:\nolinkurl{10.4324/9780367804886}}


\bibitem[Jiang et~al\mbox{.}(2012)]%
        {jiang_learning_2012}
\bibfield{author}{\bibinfo{person}{Yun Jiang}, \bibinfo{person}{Marcus Lim}, {and} \bibinfo{person}{Ashutosh Saxena}.} \bibinfo{year}{2012}\natexlab{}.
\newblock \bibinfo{title}{Learning {Object} {Arrangements} in {3D} {Scenes} using {Human} {Context}}.
\newblock
\urldef\tempurl%
\url{http://arxiv.org/abs/1206.6462}
\showURL{%
\tempurl}
\newblock
\shownote{arXiv:1206.6462}.


\bibitem[Kant et~al\mbox{.}(2022)]%
        {kant_housekeep_2022}
\bibfield{author}{\bibinfo{person}{Yash Kant}, \bibinfo{person}{Arun Ramachandran}, \bibinfo{person}{Sriram Yenamandra}, \bibinfo{person}{Igor Gilitschenski}, \bibinfo{person}{Dhruv Batra}, \bibinfo{person}{Andrew Szot}, {and} \bibinfo{person}{Harsh Agrawal}.} \bibinfo{year}{2022}\natexlab{}.
\newblock \showarticletitle{Housekeep: {Tidying} {Virtual} {Households} {Using} {Commonsense} {Reasoning}}. In \bibinfo{booktitle}{\emph{Computer {Vision} – {ECCV} 2022}}, \bibfield{editor}{\bibinfo{person}{Shai Avidan}, \bibinfo{person}{Gabriel Brostow}, \bibinfo{person}{Moustapha Cissé}, \bibinfo{person}{Giovanni~Maria Farinella}, {and} \bibinfo{person}{Tal Hassner}} (Eds.). \bibinfo{publisher}{Springer Nature Switzerland}, \bibinfo{address}{Cham}, \bibinfo{pages}{355--373}.
\newblock
\showISBNx{978-3-031-19842-7}
\href{https://doi.org/10.1007/978-3-031-19842-7_21}{doi:\nolinkurl{10.1007/978-3-031-19842-7_21}}


\bibitem[Kapelyukh and Johns(2022)]%
        {KapelyukhMyRules}
\bibfield{author}{\bibinfo{person}{Ivan Kapelyukh} {and} \bibinfo{person}{Edward Johns}.} \bibinfo{year}{2022}\natexlab{}.
\newblock \showarticletitle{My House, My Rules: Learning Tidying Preferences with Graph Neural Networks}. In \bibinfo{booktitle}{\emph{Proceedings of the 5th Conference on Robot Learning}} \emph{(\bibinfo{series}{Proceedings of Machine Learning Research}, Vol.~\bibinfo{volume}{164})}. \bibinfo{publisher}{PMLR}, \bibinfo{address}{London, UK}, \bibinfo{pages}{740--749}.
\newblock
\urldef\tempurl%
\url{https://proceedings.mlr.press/v164/kapelyukh22a.html}
\showURL{%
\tempurl}


\bibitem[Kapelyukh and Johns(2023)]%
        {kapelyukh_scenescore_2023}
\bibfield{author}{\bibinfo{person}{Ivan Kapelyukh} {and} \bibinfo{person}{Edward Johns}.} \bibinfo{year}{2023}\natexlab{}.
\newblock \bibinfo{title}{{SceneScore}: {Learning} a {Cost} {Function} for {Object} {Arrangement}}.
\newblock
\urldef\tempurl%
\url{http://arxiv.org/abs/2311.08530}
\showURL{%
\tempurl}
\newblock
\shownote{arXiv:2311.08530}.


\bibitem[Khanna et~al\mbox{.}(2024)]%
        {khanna2023habitatsyntheticscenesdataset}
\bibfield{author}{\bibinfo{person}{Mukul Khanna}, \bibinfo{person}{Yongsen Mao}, \bibinfo{person}{Hanxiao Jiang}, \bibinfo{person}{Sanjay Haresh}, \bibinfo{person}{Brennan Shacklett}, \bibinfo{person}{Dhruv Batra}, \bibinfo{person}{Alexander Clegg}, \bibinfo{person}{Eric Undersander}, \bibinfo{person}{Angel~X. Chang}, {and} \bibinfo{person}{Manolis Savva}.} \bibinfo{year}{2024}\natexlab{}.
\newblock \showarticletitle{Habitat Synthetic Scenes Dataset (HSSD-200): An Analysis of 3D Scene Scale and Realism Tradeoffs for ObjectGoal Navigation}. In \bibinfo{booktitle}{\emph{2024 IEEE/CVF Conference on Computer Vision and Pattern Recognition (CVPR)}} (Seattle, WA, USA). \bibinfo{publisher}{IEEE}, \bibinfo{address}{Piscataway, NJ, USA}, \bibinfo{pages}{16384--16393}.
\newblock
\showISBNx{979-8-3503-5300-6}
\href{https://doi.org/10.1109/CVPR52733.2024.01550}{doi:\nolinkurl{10.1109/CVPR52733.2024.01550}}


\bibitem[Kocsis and Szepesv{\'a}ri(2006)]%
        {10.1007/11871842_29}
\bibfield{author}{\bibinfo{person}{Levente Kocsis} {and} \bibinfo{person}{Csaba Szepesv{\'a}ri}.} \bibinfo{year}{2006}\natexlab{}.
\newblock \showarticletitle{Bandit Based Monte-Carlo Planning}. In \bibinfo{booktitle}{\emph{Machine Learning: ECML 2006}}, \bibfield{editor}{\bibinfo{person}{Johannes F{\"u}rnkranz}, \bibinfo{person}{Tobias Scheffer}, {and} \bibinfo{person}{Myra Spiliopoulou}} (Eds.). \bibinfo{publisher}{Springer Berlin Heidelberg}, \bibinfo{address}{Berlin, Heidelberg}, \bibinfo{pages}{282--293}.
\newblock
\showISBNx{978-3-540-46056-5}


\bibitem[Kotu and Deshpande(2019)]%
        {KOTU201965}
\bibfield{author}{\bibinfo{person}{Vijay Kotu} {and} \bibinfo{person}{Bala Deshpande}.} \bibinfo{year}{2019}\natexlab{}.
\newblock \showarticletitle{Chapter 4 - Classification}.
\newblock In \bibinfo{booktitle}{\emph{Data Science (Second Edition)} (\bibinfo{edition}{second edition} ed.)}, \bibfield{editor}{\bibinfo{person}{Vijay Kotu} {and} \bibinfo{person}{Bala Deshpande}} (Eds.). \bibinfo{publisher}{Morgan Kaufmann}, \bibinfo{address}{Burlington, MA, USA}, \bibinfo{pages}{65--163}.
\newblock
\showISBNx{978-0-12-814761-0}
\href{https://doi.org/10.1016/B978-0-12-814761-0.00004-6}{doi:\nolinkurl{10.1016/B978-0-12-814761-0.00004-6}}


\bibitem[Kroemer(2008)]%
        {kroemer_fitting_2008}
\bibfield{author}{\bibinfo{person}{Karl H.~E. Kroemer}.} \bibinfo{year}{2008}\natexlab{}.
\newblock \bibinfo{booktitle}{\emph{Fitting the {Human}: {Introduction} to {Ergonomics}, {Sixth} {Edition}} (\bibinfo{edition}{6} ed.)}.
\newblock \bibinfo{publisher}{CRC Press}, \bibinfo{address}{Boca Raton}.
\newblock
\showISBNx{978-0-429-13673-3}
\href{https://doi.org/10.1201/9781420055412}{doi:\nolinkurl{10.1201/9781420055412}}


\bibitem[Ledesma and Valero-Mora(2007)]%
        {ledesma2007determining}
\bibfield{author}{\bibinfo{person}{Rub{\'e}n~Daniel Ledesma} {and} \bibinfo{person}{Pedro Valero-Mora}.} \bibinfo{year}{2007}\natexlab{}.
\newblock \showarticletitle{Determining the Number of Factors to Retain in {EFA}: An Easy-to-Use Computer Program for Carrying Out Parallel Analysis}.
\newblock \bibinfo{journal}{\emph{Practical Assessment, Research, and Evaluation}} \bibinfo{volume}{12}, \bibinfo{number}{1}, Article \bibinfo{articleno}{2} (\bibinfo{year}{2007}), \bibinfo{numpages}{11}~pages.
\newblock
\showISSN{1531-7714}
\href{https://doi.org/10.7275/wjnc-nm63}{doi:\nolinkurl{10.7275/wjnc-nm63}}


\bibitem[Montello(1993)]%
        {montello_scale_1993}
\bibfield{author}{\bibinfo{person}{Daniel~R. Montello}.} \bibinfo{year}{1993}\natexlab{}.
\newblock \showarticletitle{Scale and multiple psychologies of space}. In \bibinfo{booktitle}{\emph{Spatial {Information} {Theory} {A} {Theoretical} {Basis} for {GIS}}}, \bibfield{editor}{\bibinfo{person}{Andrew~U. Frank} {and} \bibinfo{person}{Irene Campari}} (Eds.). \bibinfo{publisher}{Springer}, \bibinfo{address}{Berlin, Heidelberg}, \bibinfo{pages}{312--321}.
\newblock
\showISBNx{978-3-540-47966-6}
\href{https://doi.org/10.1007/3-540-57207-4_21}{doi:\nolinkurl{10.1007/3-540-57207-4_21}}


\bibitem[Neal et~al\mbox{.}(2012)]%
        {NEAL2012492}
\bibfield{author}{\bibinfo{person}{David~T. Neal}, \bibinfo{person}{Wendy Wood}, \bibinfo{person}{Jennifer~S. Labrecque}, {and} \bibinfo{person}{Phillippa Lally}.} \bibinfo{year}{2012}\natexlab{}.
\newblock \showarticletitle{How do habits guide behavior? Perceived and actual triggers of habits in daily life}.
\newblock \bibinfo{journal}{\emph{Journal of Experimental Social Psychology}} \bibinfo{volume}{48}, \bibinfo{number}{2} (\bibinfo{year}{2012}), \bibinfo{pages}{492--498}.
\newblock
\showISSN{0022-1031}
\href{https://doi.org/10.1016/j.jesp.2011.10.011}{doi:\nolinkurl{10.1016/j.jesp.2011.10.011}}


\bibitem[Newman et~al\mbox{.}(2024)]%
        {newman_degustabot_2024}
\bibfield{author}{\bibinfo{person}{Benjamin~A. Newman}, \bibinfo{person}{Pranay Gupta}, \bibinfo{person}{Kris Kitani}, \bibinfo{person}{Yonatan Bisk}, \bibinfo{person}{Henny Admoni}, {and} \bibinfo{person}{Chris Paxton}.} \bibinfo{year}{2024}\natexlab{}.
\newblock \bibinfo{title}{{DegustaBot}: {Zero}-{Shot} {Visual} {Preference} {Estimation} for {Personalized} {Multi}-{Object} {Rearrangement}}.
\newblock
\urldef\tempurl%
\url{http://arxiv.org/abs/2407.08876}
\showURL{%
\tempurl}
\newblock
\shownote{arXiv:2407.08876}.


\bibitem[O'Donnell et~al\mbox{.}(2018)]%
        {o2018semantic}
\bibfield{author}{\bibinfo{person}{Ryan~E O'Donnell}, \bibinfo{person}{Andrew Clement}, {and} \bibinfo{person}{James~R Brockmole}.} \bibinfo{year}{2018}\natexlab{}.
\newblock \showarticletitle{Semantic and functional relationships among objects increase the capacity of visual working memory.}
\newblock \bibinfo{journal}{\emph{Journal of Experimental Psychology: Learning, Memory, and Cognition}} \bibinfo{volume}{44}, \bibinfo{number}{7} (\bibinfo{year}{2018}), \bibinfo{pages}{1151}.
\newblock


\bibitem[Oliva and Torralba(2007)]%
        {oliva_role_2007}
\bibfield{author}{\bibinfo{person}{Aude Oliva} {and} \bibinfo{person}{Antonio Torralba}.} \bibinfo{year}{2007}\natexlab{}.
\newblock \showarticletitle{The role of context in object recognition}.
\newblock \bibinfo{journal}{\emph{Trends in Cognitive Sciences}} \bibinfo{volume}{11}, \bibinfo{number}{12} (\bibinfo{date}{Dec.} \bibinfo{year}{2007}), \bibinfo{pages}{520--527}.
\newblock
\showISSN{1364-6613, 1879-307X}
\href{https://doi.org/10.1016/j.tics.2007.09.009}{doi:\nolinkurl{10.1016/j.tics.2007.09.009}}
\newblock
\shownote{Publisher: Elsevier}.


\bibitem[Paxton et~al\mbox{.}(2022)]%
        {paxton2022predicting}
\bibfield{author}{\bibinfo{person}{Chris Paxton}, \bibinfo{person}{Chris Xie}, \bibinfo{person}{Tucker Hermans}, {and} \bibinfo{person}{Dieter Fox}.} \bibinfo{year}{2022}\natexlab{}.
\newblock \showarticletitle{Predicting Stable Configurations for Semantic Placement of Novel Objects}. In \bibinfo{booktitle}{\emph{Proceedings of the 5th Conference on Robot Learning}} \emph{(\bibinfo{series}{Proceedings of Machine Learning Research}, Vol.~\bibinfo{volume}{164})}, \bibfield{editor}{\bibinfo{person}{Aleksandra Faust}, \bibinfo{person}{David Hsu}, {and} \bibinfo{person}{Gerhard Neumann}} (Eds.). \bibinfo{publisher}{PMLR}, \bibinfo{address}{London, UK}, \bibinfo{pages}{806--815}.
\newblock
\urldef\tempurl%
\url{https://proceedings.mlr.press/v164/paxton22a.html}
\showURL{%
\tempurl}


\bibitem[Pheasant and Haslegrave(2018)]%
        {pheasant_bodyspace_2018}
\bibfield{author}{\bibinfo{person}{Stephen Pheasant} {and} \bibinfo{person}{Christine~M. Haslegrave}.} \bibinfo{year}{2018}\natexlab{}.
\newblock \bibinfo{booktitle}{\emph{Bodyspace: {Anthropometry}, {Ergonomics} and the {Design} of {Work}, {Third} {Edition}} (\bibinfo{edition}{3} ed.)}.
\newblock \bibinfo{publisher}{CRC Press}, \bibinfo{address}{Boca Raton}.
\newblock
\showISBNx{978-1-315-37521-2}
\href{https://doi.org/10.1201/9781315375212}{doi:\nolinkurl{10.1201/9781315375212}}


\bibitem[{R Core Team}(2025)]%
        {Rcore2025}
\bibfield{author}{\bibinfo{person}{{R Core Team}}.} \bibinfo{year}{2025}\natexlab{}.
\newblock \bibinfo{booktitle}{\emph{R: A Language and Environment for Statistical Computing (Version 4.5)}}.
\newblock R Foundation for Statistical Computing.
\newblock
\urldef\tempurl%
\url{https://cran.r-project.org}
\showURL{%
\tempurl}


\bibitem[Ramachandruni and Chernova(2025)]%
        {ramachandruni2025personalizedroboticobjectrearrangement}
\bibfield{author}{\bibinfo{person}{Kartik Ramachandruni} {and} \bibinfo{person}{Sonia Chernova}.} \bibinfo{year}{2025}\natexlab{}.
\newblock \bibinfo{title}{Personalized Robotic Object Rearrangement from Scene Context}.
\newblock
\showeprint[arxiv]{2505.11108}~[cs.RO]
\urldef\tempurl%
\url{https://arxiv.org/abs/2505.11108}
\showURL{%
\tempurl}


\bibitem[Ramachandruni et~al\mbox{.}(2023)]%
        {CONSOR_chernova}
\bibfield{author}{\bibinfo{person}{Kartik Ramachandruni}, \bibinfo{person}{Max Zuo}, {and} \bibinfo{person}{Sonia Chernova}.} \bibinfo{year}{2023}\natexlab{}.
\newblock \showarticletitle{ConSOR: A Context-Aware Semantic Object Rearrangement Framework for Partially Arranged Scenes}. In \bibinfo{booktitle}{\emph{2023 IEEE/RSJ International Conference on Intelligent Robots and Systems (IROS)}}. \bibinfo{publisher}{IEEE}, \bibinfo{address}{Detroit, MI, USA}, \bibinfo{pages}{82--89}.
\newblock
\href{https://doi.org/10.1109/IROS55552.2023.10341873}{doi:\nolinkurl{10.1109/IROS55552.2023.10341873}}


\bibitem[Rodríguez-Lera et~al\mbox{.}(2024)]%
        {rodríguezlera2024roxiedefiningroboticexplanation}
\bibfield{author}{\bibinfo{person}{Francisco~J. Rodríguez-Lera}, \bibinfo{person}{Miguel~A. González-Santamarta}, \bibinfo{person}{Alejandro González-Cantón}, \bibinfo{person}{Laura Fernández-Becerra}, \bibinfo{person}{David Sobrín-Hidalgo}, {and} \bibinfo{person}{Angel~Manuel Guerrero-Higueras}.} \bibinfo{year}{2024}\natexlab{}.
\newblock \bibinfo{title}{ROXIE: Defining a Robotic eXplanation and Interpretability Engine}.
\newblock
\showeprint[arxiv]{2403.16606}~[cs.RO]
\urldef\tempurl%
\url{https://arxiv.org/abs/2403.16606}
\showURL{%
\tempurl}


\bibitem[Rudin(2019)]%
        {rudin2019stopexplainingblackbox}
\bibfield{author}{\bibinfo{person}{Cynthia Rudin}.} \bibinfo{year}{2019}\natexlab{}.
\newblock \showarticletitle{Stop explaining black box machine learning models for high stakes decisions and use interpretable models instead}.
\newblock \bibinfo{journal}{\emph{Nature Machine Intelligence}} \bibinfo{volume}{1}, \bibinfo{number}{5} (\bibinfo{date}{May} \bibinfo{year}{2019}), \bibinfo{pages}{206--215}.
\newblock
\showISSN{2522-5839}
\href{https://doi.org/10.1038/s42256-019-0048-x}{doi:\nolinkurl{10.1038/s42256-019-0048-x}}
\newblock
\shownote{Publisher: Nature Publishing Group}.


\bibitem[Sakai and Nagai(2022)]%
        {sakai2021explainableautonomousrobotssurvey}
\bibfield{author}{\bibinfo{person}{Tatsuya Sakai} {and} \bibinfo{person}{Takayuki Nagai}.} \bibinfo{year}{2022}\natexlab{}.
\newblock \showarticletitle{Explainable autonomous robots: a survey and perspective}.
\newblock \bibinfo{journal}{\emph{Advanced Robotics}} \bibinfo{volume}{36}, \bibinfo{number}{5-6} (\bibinfo{date}{March} \bibinfo{year}{2022}), \bibinfo{pages}{219--238}.
\newblock
\showISSN{0169-1864}
\href{https://doi.org/10.1080/01691864.2022.2029720}{doi:\nolinkurl{10.1080/01691864.2022.2029720}}
\newblock
\shownote{Publisher: Taylor \& Francis \_eprint: https://doi.org/10.1080/01691864.2022.2029720}.


\bibitem[Sarch et~al\mbox{.}(2022)]%
        {sarch_tidee_2022}
\bibfield{author}{\bibinfo{person}{Gabriel Sarch}, \bibinfo{person}{Zhaoyuan Fang}, \bibinfo{person}{Adam~W. Harley}, \bibinfo{person}{Paul Schydlo}, \bibinfo{person}{Michael~J. Tarr}, \bibinfo{person}{Saurabh Gupta}, {and} \bibinfo{person}{Katerina Fragkiadaki}.} \bibinfo{year}{2022}\natexlab{}.
\newblock \showarticletitle{{TIDEE}: {Tidying} {Up} {Novel} {Rooms} {Using} {Visuo}-{Semantic} {Commonsense} {Priors}}. In \bibinfo{booktitle}{\emph{Computer {Vision} – {ECCV} 2022: 17th {European} {Conference}, {Tel} {Aviv}, {Israel}, {October} 23–27, 2022, {Proceedings}, {Part} {XXXIX}}}. \bibinfo{publisher}{Springer-Verlag}, \bibinfo{address}{Berlin, Heidelberg}, \bibinfo{pages}{480--496}.
\newblock
\showISBNx{978-3-031-19841-0}
\href{https://doi.org/10.1007/978-3-031-19842-7_28}{doi:\nolinkurl{10.1007/978-3-031-19842-7_28}}


\bibitem[Taber(2018)]%
        {taber2018use}
\bibfield{author}{\bibinfo{person}{Keith~S Taber}.} \bibinfo{year}{2018}\natexlab{}.
\newblock \showarticletitle{The use of Cronbach’s alpha when developing and reporting research instruments in science education}.
\newblock \bibinfo{journal}{\emph{Research in science education}} \bibinfo{volume}{48}, \bibinfo{number}{6} (\bibinfo{year}{2018}), \bibinfo{pages}{1273--1296}.
\newblock


\bibitem[{The jamovi project}(2025)]%
        {jamovi2025}
\bibfield{author}{\bibinfo{person}{{The jamovi project}}.} \bibinfo{year}{2025}\natexlab{}.
\newblock \bibinfo{booktitle}{\emph{jamovi (Version 2.7) [Computer software]}}.
\newblock
\urldef\tempurl%
\url{https://www.jamovi.org}
\showURL{%
\tempurl}


\bibitem[{The Royal Children's Hospital Melbourne}(2025)]%
        {rch_poisoning_2025}
\bibfield{author}{\bibinfo{person}{{The Royal Children's Hospital Melbourne}}.} \bibinfo{year}{2025}\natexlab{}.
\newblock \bibinfo{title}{Safety: Poisoning prevention}.
\newblock \bibinfo{howpublished}{\url{https://www.rch.org.au/kidsinfo/fact_sheets/safety_poisoning_prevention/}}.
\newblock
\newblock
\shownote{Reviewed October 2018; Updated July 2025. Advice: store chemicals/cleaners/medicines in locked or child-resistant cupboards, out of reach and out of sight ($\geq$1.5 m). Accessed: 2025-10-01.}.


\bibitem[Trabucco et~al\mbox{.}(2022)]%
        {trabucco2022simpleapproachvisualrearrangement}
\bibfield{author}{\bibinfo{person}{Brandon Trabucco}, \bibinfo{person}{Gunnar Sigurdsson}, \bibinfo{person}{Robinson Piramuthu}, \bibinfo{person}{Gaurav~S. Sukhatme}, {and} \bibinfo{person}{Ruslan Salakhutdinov}.} \bibinfo{year}{2022}\natexlab{}.
\newblock \bibinfo{title}{A Simple Approach for Visual Rearrangement: 3D Mapping and Semantic Search}.
\newblock
\showeprint[arxiv]{2206.13396}~[cs.CV]
\urldef\tempurl%
\url{https://arxiv.org/abs/2206.13396}
\showURL{%
\tempurl}


\bibitem[{U.S. Food and Drug Administration}(2022)]%
        {fda_foodcode_2022}
\bibfield{author}{\bibinfo{person}{{U.S. Food and Drug Administration}}.} \bibinfo{year}{2022}\natexlab{}.
\newblock \bibinfo{booktitle}{\emph{Food Code 2022: Model Food Code for Retail and Food Service}}.
\newblock \bibinfo{type}{{T}echnical {R}eport}. \bibinfo{institution}{U.S. Department of Health and Human Services / FDA}.
\newblock
\urldef\tempurl%
\url{https://www.fda.gov/food/fda-food-code/food-code-2022}
\showURL{%
\tempurl}
\newblock
\shownote{Model code for retail and food service food safety; accessed 2025-10-01.}.


\bibitem[V{\~o} et~al\mbox{.}(2019)]%
        {vo2019reading}
\bibfield{author}{\bibinfo{person}{Melissa Le-Hoa V{\~o}}, \bibinfo{person}{Sage~EP Boettcher}, {and} \bibinfo{person}{Dejan Draschkow}.} \bibinfo{year}{2019}\natexlab{}.
\newblock \showarticletitle{Reading scenes: How scene grammar guides attention and aids perception in real-world environments}.
\newblock \bibinfo{journal}{\emph{Current opinion in psychology}}  \bibinfo{volume}{29} (\bibinfo{year}{2019}), \bibinfo{pages}{205--210}.
\newblock


\bibitem[Võ and Wolfe(2013a)]%
        {vo_differential_2013}
\bibfield{author}{\bibinfo{person}{Melissa L.-H. Võ} {and} \bibinfo{person}{Jeremy~M. Wolfe}.} \bibinfo{year}{2013}\natexlab{a}.
\newblock \showarticletitle{Differential {Electrophysiological} {Signatures} of {Semantic} and {Syntactic} {Scene} {Processing}}.
\newblock \bibinfo{journal}{\emph{Psychological Science}} \bibinfo{volume}{24}, \bibinfo{number}{9} (\bibinfo{date}{Sept.} \bibinfo{year}{2013}), \bibinfo{pages}{1816--1823}.
\newblock
\showISSN{0956-7976}
\href{https://doi.org/10.1177/0956797613476955}{doi:\nolinkurl{10.1177/0956797613476955}}
\newblock
\shownote{Publisher: SAGE Publications Inc}.


\bibitem[Võ and Wolfe(2013b)]%
        {doi:10.1177/0956797613476955}
\bibfield{author}{\bibinfo{person}{Melissa L.-H. Võ} {and} \bibinfo{person}{Jeremy~M. Wolfe}.} \bibinfo{year}{2013}\natexlab{b}.
\newblock \showarticletitle{Differential Electrophysiological Signatures of Semantic and Syntactic Scene Processing}.
\newblock \bibinfo{journal}{\emph{Psychological Science}} \bibinfo{volume}{24}, \bibinfo{number}{9} (\bibinfo{year}{2013}), \bibinfo{pages}{1816--1823}.
\newblock
\showeprint{https://doi.org/10.1177/0956797613476955}
\href{https://doi.org/10.1177/0956797613476955}{doi:\nolinkurl{10.1177/0956797613476955}}
\newblock
\shownote{PMID: 23842954}.


\bibitem[Wang et~al\mbox{.}(2025)]%
        {wang2024apricotactivepreferencelearning}
\bibfield{author}{\bibinfo{person}{Huaxiaoyue Wang}, \bibinfo{person}{Nathaniel Chin}, \bibinfo{person}{Gonzalo Gonzalez-Pumariega}, \bibinfo{person}{Xiangwan Sun}, \bibinfo{person}{Neha Sunkara}, \bibinfo{person}{Maximus~Adrian Pace}, \bibinfo{person}{Jeannette Bohg}, {and} \bibinfo{person}{Sanjiban Choudhury}.} \bibinfo{year}{2025}\natexlab{}.
\newblock \showarticletitle{APRICOT: Active Preference Learning and Constraint-Aware Task Planning with LLMs}. In \bibinfo{booktitle}{\emph{Proceedings of The 8th Conference on Robot Learning}} \emph{(\bibinfo{series}{Proceedings of Machine Learning Research}, Vol.~\bibinfo{volume}{270})}. \bibinfo{publisher}{PMLR}, \bibinfo{address}{Munich, Germany}, \bibinfo{pages}{1590--1642}.
\newblock
\urldef\tempurl%
\url{https://proceedings.mlr.press/v270/wang25e.html}
\showURL{%
\tempurl}


\bibitem[Wood and R{\"u}nger(2016)]%
        {wood2016psychology}
\bibfield{author}{\bibinfo{person}{Wendy Wood} {and} \bibinfo{person}{Dennis R{\"u}nger}.} \bibinfo{year}{2016}\natexlab{}.
\newblock \showarticletitle{Psychology of habit}.
\newblock \bibinfo{journal}{\emph{Annual review of psychology}} \bibinfo{volume}{67}, \bibinfo{number}{1} (\bibinfo{year}{2016}), \bibinfo{pages}{289--314}.
\newblock


\bibitem[Wu et~al\mbox{.}(2023)]%
        {wu_tidybot_2023}
\bibfield{author}{\bibinfo{person}{Jimmy Wu}, \bibinfo{person}{Rika Antonova}, \bibinfo{person}{Adam Kan}, \bibinfo{person}{Marion Lepert}, \bibinfo{person}{Andy Zeng}, \bibinfo{person}{Shuran Song}, \bibinfo{person}{Jeannette Bohg}, \bibinfo{person}{Szymon Rusinkiewicz}, {and} \bibinfo{person}{Thomas Funkhouser}.} \bibinfo{year}{2023}\natexlab{}.
\newblock \showarticletitle{{TidyBot}: personalized robot assistance with large language models}.
\newblock \bibinfo{journal}{\emph{Autonomous Robots}} \bibinfo{volume}{47}, \bibinfo{number}{8} (\bibinfo{date}{Dec.} \bibinfo{year}{2023}), \bibinfo{pages}{1087--1102}.
\newblock
\showISSN{1573-7527}
\href{https://doi.org/10.1007/s10514-023-10139-z}{doi:\nolinkurl{10.1007/s10514-023-10139-z}}


\bibitem[Yang and Abdel-Malek(2009)]%
        {yang2009human}
\bibfield{author}{\bibinfo{person}{Jingzhou Yang} {and} \bibinfo{person}{Karim Abdel-Malek}.} \bibinfo{year}{2009}\natexlab{}.
\newblock \showarticletitle{Human reach envelope and zone differentiation for ergonomic design}.
\newblock \bibinfo{journal}{\emph{Human Factors and Ergonomics in Manufacturing \& Service Industries}} \bibinfo{volume}{19}, \bibinfo{number}{1} (\bibinfo{year}{2009}), \bibinfo{pages}{15--34}.
\newblock


\bibitem[Świechowski et~al\mbox{.}(2023)]%
        {swiechowski_monte_2023}
\bibfield{author}{\bibinfo{person}{Maciej Świechowski}, \bibinfo{person}{Konrad Godlewski}, \bibinfo{person}{Bartosz Sawicki}, {and} \bibinfo{person}{Jacek Mańdziuk}.} \bibinfo{year}{2023}\natexlab{}.
\newblock \showarticletitle{Monte {Carlo} {Tree} {Search}: a review of recent modifications and applications}.
\newblock \bibinfo{journal}{\emph{Artificial Intelligence Review}} \bibinfo{volume}{56}, \bibinfo{number}{3} (\bibinfo{date}{March} \bibinfo{year}{2023}), \bibinfo{pages}{2497--2562}.
\newblock
\showISSN{1573-7462}
\href{https://doi.org/10.1007/s10462-022-10228-y}{doi:\nolinkurl{10.1007/s10462-022-10228-y}}


\end{thebibliography}

\end{document}


\maketitle

\bigskip
\section*{Appendix contents}
This appendix includes: (A) the Living Room survey interface and (B) representative case studies showing predicted vs.\ ground-truth placements for four example participant weightings.

\bigskip
\section{Survey Interface (Living Room)}

\begin{figure}[h]
  \centering
  \includegraphics[width=\textwidth,height=0.5\textheight,keepaspectratio]{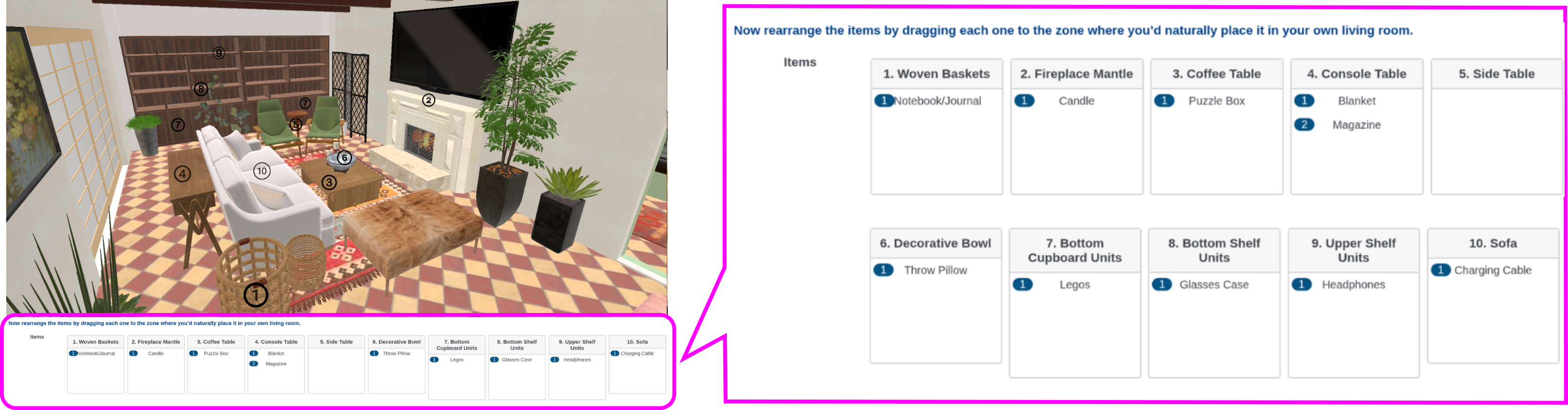}
  \caption{Survey interface showing the Living Room organization task (omitted from Fig.~1 in the main paper due to space). Participants dragged objects into receptacle zones.}
  \label{fig:interface_livingroom}
\end{figure}

\clearpage
\section{Representative Case Studies (Supplement to Section 4.3: Preference-Aligned Trajectory Generation)}

\begin{tcolorbox}[casebox style=blue]
\textbf{\large Spatial-dominant (Living Room, P23)}\\[4pt]
\CaseHeaderLines{P23}{Living Room}{0.37, 0.29, 0.17, 0.17}
\textbf{Per-object accuracy:} 60\%

\vspace{2mm}
\begin{tabularx}{\linewidth}{@{} L C C >{\centering\arraybackslash}p{0.9cm} @{}}
\toprule
\textbf{Item} & \textbf{Predicted receptacle} & \textbf{Participant (GT) receptacle} & \textbf{OK} \\
\midrule
Blanket & sofa & sofa & \cmark \\
Candle & fireplace mantle & fireplace mantle & \cmark \\
Charging cable & coffee table & coffee table & \cmark \\
Glasses case & coffee table & coffee table & \cmark \\
Headphones & bottom cupboard units & console table & \xmark \\
Legos & sofa & bottom cupboard units & \xmark \\
Magazine & side table & coffee table & \xmark \\
Notebook / Journal & bottom shelf units & bottom shelf units & \cmark \\
Puzzle box & bottom cupboard units & bottom cupboard units & \cmark \\
Throw pillow & decorative bowl & sofa & \xmark \\
\bottomrule
\end{tabularx}
\end{tcolorbox}

\vspace{6mm}

\begin{tcolorbox}[casebox style=teal]
\textbf{\large Balanced (Living Room, P32)}\\[4pt]
\CaseHeaderLines{P32}{Living Room}{0.25, 0.25, 0.25, 0.25}
\textbf{Per-object accuracy:} 40\%

\vspace{2mm}
\begin{tabularx}{\linewidth}{@{} L C C >{\centering\arraybackslash}p{0.9cm} @{}}
\toprule
\textbf{Item} & \textbf{Predicted receptacle} & \textbf{Participant (GT) receptacle} & \textbf{OK} \\
\midrule
Blanket & bottom cupboard units & bottom cupboard units & \cmark \\
Candle & fireplace mantle & fireplace mantle & \cmark \\
Charging cable & coffee table & bottom shelf units & \xmark \\
Glasses case & coffee table & coffee table & \cmark \\
Headphones & side table & coffee table & \xmark \\
Legos & bottom cupboard units & console table & \xmark \\
Magazine & side table & console table & \xmark \\
Notebook / Journal & bottom shelf units & side table & \xmark \\
Puzzle box & console table & console table & \cmark \\
Throw pillow & decorative bowl & sofa & \xmark \\
\bottomrule
\end{tabularx}
\end{tcolorbox}

\vspace{6mm}

\begin{tcolorbox}[casebox style=orange]
\textbf{\large Habitual-dominant (Kitchen, P24)}\\[4pt]
\CaseHeaderLines{P24}{Kitchen}{0.34, 0.40, 0.18, 0.08}
\textbf{Per-object accuracy:} 80\%

\vspace{2mm}
\begin{tabularx}{\linewidth}{@{} L C C >{\centering\arraybackslash}p{0.9cm} @{}}
\toprule
\textbf{Item} & \textbf{Predicted receptacle} & \textbf{Participant (GT) receptacle} & \textbf{OK} \\
\midrule
Apples & fridge interior & fridge interior & \cmark \\
Bag of potatoes & black metal rack & fridge interior & \xmark \\
Bananas & fridge interior & fridge interior & \cmark \\
Cracker box & fridge interior & kitchen island & \xmark \\
Cutlery & drawers & drawers & \cmark \\
Electric kettle & countertop & countertop & \cmark \\
Reusable shopping bags & black metal rack & black metal rack & \cmark \\
Spice jar & upper cupboards & upper cupboards & \cmark \\
Surface cleaner bottle & lower cabinet & lower cabinet & \cmark \\
Tea towel & lower cabinet & lower cabinet & \cmark \\
\bottomrule
\end{tabularx}
\end{tcolorbox}

\vspace{6mm}

\begin{tcolorbox}[casebox style=purple]
\textbf{\large Semantic-dominant (Kitchen, P16)}\\[4pt]
\CaseHeaderLines{P16}{Kitchen}{0.26, 0.21, 0.30, 0.23}
\textbf{Per-object accuracy:} 90\%

\vspace{2mm}
\begin{tabularx}{\linewidth}{@{} L C C >{\centering\arraybackslash}p{0.9cm} @{}}
\toprule
\textbf{Item} & \textbf{Predicted receptacle} & \textbf{Participant (GT) receptacle} & \textbf{OK} \\
\midrule
Apples & bowl on kitchen island & bowl on kitchen island & \cmark \\
Bag of potatoes & black metal rack & lower cabinet & \xmark \\
Bananas & bowl on kitchen island & bowl on kitchen island & \cmark \\
Cracker box & countertop & countertop & \cmark \\
Cutlery & drawers & drawers & \cmark \\
Electric kettle & countertop & countertop & \cmark \\
Reusable shopping bags & lower cabinet & lower cabinet & \cmark \\
Spice jar & upper cupboards & upper cupboards & \cmark \\
Surface cleaner bottle & lower cabinet & lower cabinet & \cmark \\
Tea towel & oven handle bar & oven handle bar & \cmark \\
\bottomrule
\end{tabularx}
\end{tcolorbox}

\bigskip
\noindent\textit{Note.} Each case shows a representative participant profile (ID), the normalized weight vector across constructs (S = spatial, H = habitual, Se = semantic, C = commonsense), and per-object accuracy computed as the proportion of objects placed in the same receptacle as the participant (GT).